\pdfoutput=1

\documentclass[11pt]{article}

\usepackage[final]{acl}

\usepackage{times}
\usepackage{latexsym}

\usepackage[T1]{fontenc}

\usepackage[utf8]{inputenc}

\usepackage{microtype}

\usepackage{inconsolata}

\usepackage{graphicx}

\usepackage{amsmath}
\usepackage{booktabs}
\usepackage{array}
\usepackage{graphicx}
\usepackage{caption} 
\usepackage{makecell} 
\usepackage{multirow}
\usepackage{enumitem}
\usepackage{siunitx}

\title{Dialogue Systems for Emotional Support via Value Reinforcement}

\author{
Juhee Kim\textsuperscript{1}\quad
Chunghu Mok\textsuperscript{2}\quad
Jisun Lee\textsuperscript{2}\quad
Hyang Sook Kim\textsuperscript{2}\quad
Yohan Jo\textsuperscript{1}\thanks{Corresponding author}\
\\  
\textsuperscript{1}Graduate School of Data Science,
\textsuperscript{2}Department of Psychology
\\
Seoul National University
\\
\texttt{\{inbz1244, chunghumok, jisune87, hyangkim, yohan.jo\}@snu.ac.kr}
}

\begin{document}
\maketitle
\begin{abstract}
Emotional support dialogue systems aim to reduce help-seekers' distress and help them overcome challenges. 
While human values---core beliefs that shape an individual's priorities---are increasingly emphasized in contemporary psychological therapy for their role in fostering internal transformation and long-term emotional well-being, their integration into emotional support systems remains underexplored. 
To bridge this gap, we present a value-driven method for training emotional support dialogue systems designed to reinforce positive \textit{values} in seekers. 
Notably, our model identifies which values to reinforce at each turn and how to do so, by leveraging online support conversations from Reddit.
We evaluate the method across support skills, seekers' emotional intensity, and value reinforcement.
Our method consistently outperforms various baselines, effectively exploring and eliciting values from seekers. Additionally, leveraging crowd knowledge from Reddit significantly enhances its effectiveness.
Therapists highlighted its ability to validate seekers' challenges and emphasize positive aspects of their situations---both crucial elements of value reinforcement. 
Our work, being the first to integrate value reinforcement into emotional support systems, demonstrates its promise and establishes a foundation for future research.\footnote{Our code and dataset are available at \url{https://github.com/holi-lab/ES-Value}.}
\end{abstract}

\begin{figure}[!t]
    \includegraphics[width=\columnwidth]{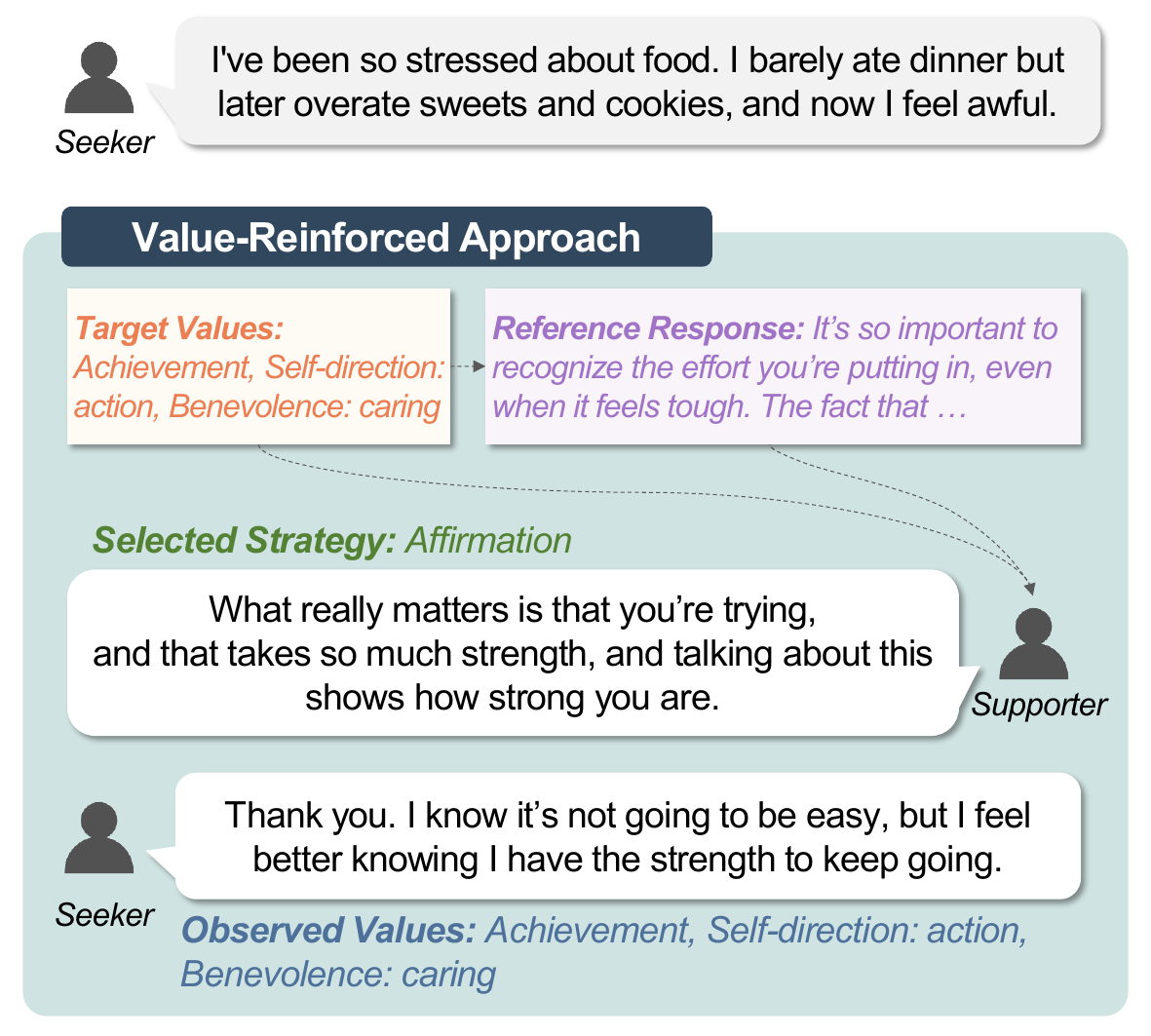}  
    \caption{An example dialogue based on our method. The supporter model receives the target values to reinforce from the seeker and a reference response for guidance. It then selects an appropriate emotional support strategy and generates a response for the next turn.}
    \label{figure_value-reinforced-example} 
\end{figure}

\section{Introduction}
Emotional support aims to help individuals (\textit{seekers}) in addressing everyday emotional difficulties, such as relationship conflicts and workplace stress, by offering reassurance, acceptance, and encouragement \citep{atoum2018emotional, burleson2003emotional}. Recent advancements in large language models have accelerated the development of dialogue systems designed to provide emotional support (\textit{supporters}) \citep{deng2024ppdpp, zhang-etal-2023-ask, chen2023kesconvknowledgeinjectionemotional}. Many models have focused on reinforcing positive emotions in seekers. However, emotional changes alone may not adequately capture deeper intrinsic transformations within the seeker, potentially reducing the long-term impact of emotional support \citep{blackledge2001emotion}.
For instance, a seeker's perfunctory \textit{``Thank you''}, used as a conversational pleasantry, receives a higher positivity score (0.758) from a sentiment classifier than the response shown in Figure~\ref{figure_value-reinforced-example} (0.583).\footnote{We employed EmoLlama-Chat-7B \citep{liu2024emollms}, which demonstrates superior performance in this task.} The low positivity score of the latter response may be attributed to phrases like \textit{``I know it’s not going to be easy''}, which could be perceived as negative. However, the latter response demonstrates a stronger commitment and willingness to change, highlighting the importance of evaluating support effectiveness beyond emotion alone.

To overcome these limitations, we propose an emotional support approach grounded in \textbf{value reinforcement}. Human values, which represent core beliefs and guiding principles, help individuals determine what is important and meaningful in life \citep{searle2003rationality}, such as self-direction, benevolence, tradition, etc.
Given their deep connection to life purpose and personal identity, values play a central role in modern psychological interventions, such as \textit{Acceptance and Commitment Therapy (ACT)} \citep{plumb2009search, hayes2005acceptance} and \textit{Values Affirmation Interventions} \citep{miyake2010reducing, jordt2017values}. 
These techniques aim to help seekers commit to goals aligned with their values, fostering intrinsic and long-term transformation. This supports the ultimate goal of achieving a healthy life---not merely feeling good but living well.
The importance of values in emotional support is further demonstrated by the widely used emotional support dataset, ESConv \citep{liu-etal-2021-towards}. Our analysis reveals that positive values are more prominently expressed in the high-effectiveness group of seekers (i.e., high reduction in negative emotions) (see Section~\ref{sec:analysis-esconv} for details).

In this paper, we present a framework for training a supporter model through simulations with a seeker simulator. To enhance the supporter’s ability to reinforce the seeker’s values, we introduce two key components trained on Reddit data: (1) a \textbf{target value detector} that identifies the values to promote at each turn, and (2) a \textbf{reference generator} that generates a supporter response to reinforce these values. 
By integrating their outputs, the supporter model aims to maximize the reward of value promotion reflected in the seeker's responses.
Figure~\ref{figure_value-reinforced-example} illustrates our approach applied to an example dialogue, which reinforces the seeker's values along with their acceptance of support and willingness to change.
The training involves two phases: supervised fine-tuning, which distills the simulation capability of GPT-4o-mini into a smaller model, and direct policy optimization \citep{NEURIPS2023_dpo}, which enhances the model's value reinforcement effectiveness.

We conducted a comprehensive evaluation in terms of supporter capabilities, the seeker's ultimate relief, and value reinforcement. The results demonstrate that our model outperforms most baselines in supporter capabilities and value reinforcement, while maintaining a competitive level of seeker relief. Notably, the model's strength in value reinforcement is highlighted in evaluations by expert therapists.\footnote{All therapists mentioned in this paper refer to two licensed clinical psychologists with over three years of clinical experience, who are also co-authors of this paper.} Specifically, it excels at effectively validating the seeker's challenges and emphasizing positive aspects of the seeker's situation, which form the foundation of value reinforcement.
These results highlight that value reinforcement is a promising direction for future research.

Below is a summary of our key contributions:
\begin{itemize}[topsep=3pt, partopsep=0pt, itemsep=3pt, parsep=0pt]
    \item To the best of our knowledge, this is the first work to explicitly integrate value reinforcement into emotional support systems.
    \item We propose an effective two-phase approach, featuring a target value detector and a reference generator, both trained on real-world knowledge from Reddit.
    \item Our approach achieves significant improvements in emotional relief and value reinforcement, paving the way for incorporating values into emotional support systems.
\end{itemize}

\section{Related Work}

\subsection{Human Values in Emotional Support}

Human values are fundamental beliefs that help individuals identify what is important and worth pursuing in life \citep{searle2003rationality}. Making decisions aligned with one's values enhances psychological flexibility---the ability to adapt effectively to life’s challenges \citep{hayes2006acceptance}---and supports long-term outcomes, such as academic achievement \citep{cohen2006reducing, cohen2009recursive}. Furthermore, value reinforcement can strengthen the connection between seekers and supporters, establishing a foundation for more effective and supportive conversations \citep{wilson2004values}. By encouraging seekers to connect with and act on their values, value reinforcement fosters long-term positive changes and enriches interpersonal dynamics, making conversations more meaningful and impactful.

\subsection{Dialogue Systems for Emotional Support}

To enhance supporter models, researchers have explored various approaches. One method uses large language models to generate diverse conversations for supporter model training \citep{zheng-etal-2024-self, liu2023chatcounselor, qiu-etal-2024-smile}. Other studies predict seekers' future states to refine supporter model training \citep{zhou-etal-2023-facilitating, cheng-etal-2022-improving, 9054379}. Recent efforts also leverage multi-turn simulations with seeker simulators to predict future responses \citep{deng2024ppdpp}.
However, most studies often overlook the role of human values. While our study builds on simulation-based training, its main contribution lies in integrating values into emotional support, emphasizing their critical role in improving system effectiveness.

\section{Value Effects in Emotional Support}
\label{sec:analysis-esconv}

This section explores the significance of value reinforcement in effective emotional support, providing the foundation for our research.

\subsection{Taxonomy for Human Values}
\label{sec:value-taxonomy}

In this study, we adopt the value taxonomy introduced by \citet{kiesel2022identifying}, which integrates the \textit{Schwartz Theory of Basic Values} (\citealp{schwartz2012refining}) with three other major value lists \citep{rokeach1973nature, brown2002life, haerpfer2020world}. The \textit{Schwartz Theory of Basic Values} has been extensively used in prior research across both NLP \citep{Kang.2023,yao-etal-2024-value, van-der-meer-etal-2023-differences, kiesel-etal-2023-semeval} and the social sciences, including the \textit{European Social Survey (ESS)}, which is designed to track changes in people's attitudes, beliefs, and behavior patterns across European nations \citep{davidov2008bringing}. This integrated taxonomy encompasses a comprehensive range of human values, organizing them into 20 value categories.
Further details on these values can be found in Table~\ref{table:value-taxonomy}.

\subsection{Exploring the Impact of Values on Emotional Support Effectiveness}

To motivate our research, we conducted an analysis to examine the role of values in emotional support by analyzing the ESConv dataset \citep{liu-etal-2021-towards}, which contains multi-turn emotional support conversations in English among crowdworkers.
We analyze whether reinforcing a seeker's values positively influences the effectiveness of emotional support.

\paragraph{Method.}
In ESConv, seekers rated the intensity of their negative emotions before and after the conversation on a scale from 1 (lowest) to 5 (highest).
In our analysis, dialogues with an initial intensity of 5 are divided into two groups: \textit{high effectiveness} (final intensity of 1--2) and \textit{low effectiveness} (final intensity of 3--4)\footnote{There were no cases in the ESConv where the final emotional intensity remained at 5.}. We then analyze positive value expressions in the seekers' final four turns using automated classifiers \citep{liu2024emollms, schroter-etal-2023-adam}.
This focus on the final four turns accounts for differences in turn length across groups and captures changes resulting from the emotional support conversation.
Detailed experimental procedures are described in Appendix~\ref{sec:appendix-value-effect}.

\begin{figure}[!t]
    \includegraphics[width=\columnwidth]{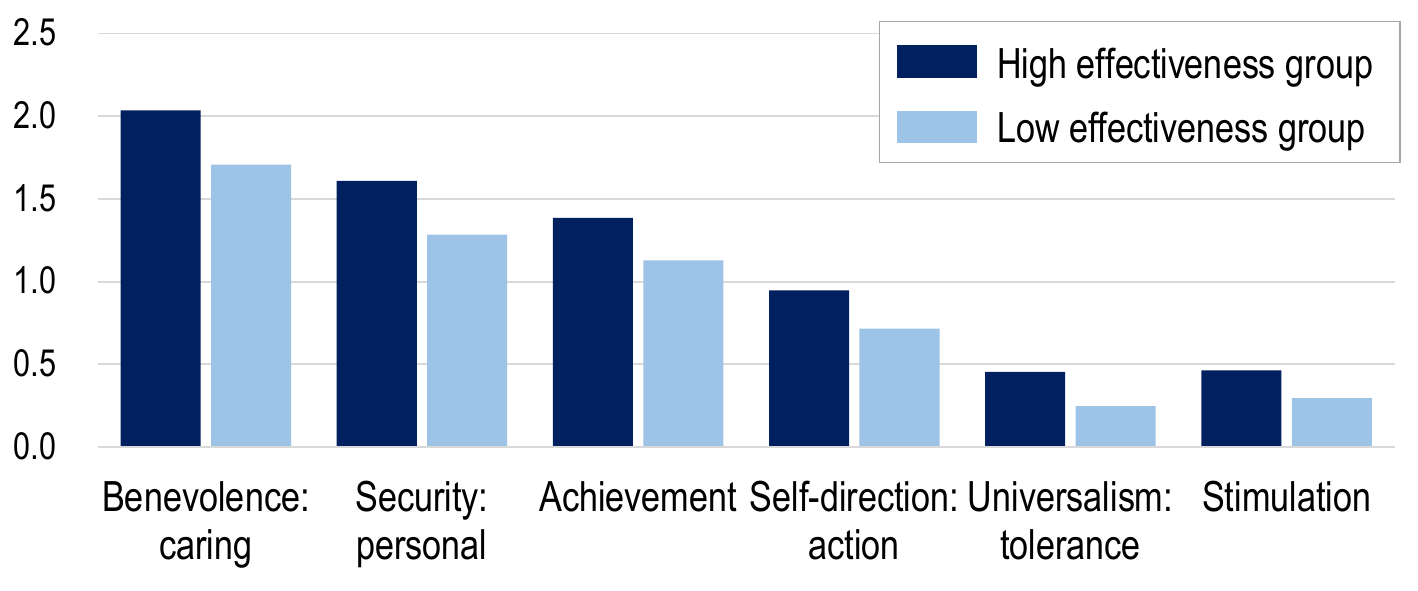}  
    \caption{Average number of value expressions in the last four turns in ESConv for high and low effectiveness groups.}
    \label{fig:esconv-analysis} 
\end{figure}

\paragraph{Results.}
According to the analysis, the \textit{high effectiveness group} exhibited a significantly higher average number of positive values (7.9) than the \textit{low effectiveness group} (6.5) in the last four turns. Figure~\ref{fig:esconv-analysis} highlights values that were pronounced in the \textit{high effectiveness group}. 
Table~\ref{table:esconv-example} provides examples of seekers' utterances that illustrate these values.

These findings support that value reinforcement in seekers positively impacts the effectiveness of emotional support and motivate our research approach to designing dialogue systems that aim to reinforce seekers' values.

\begin{figure*}[!t]
    \centering
    \includegraphics[width=\textwidth]{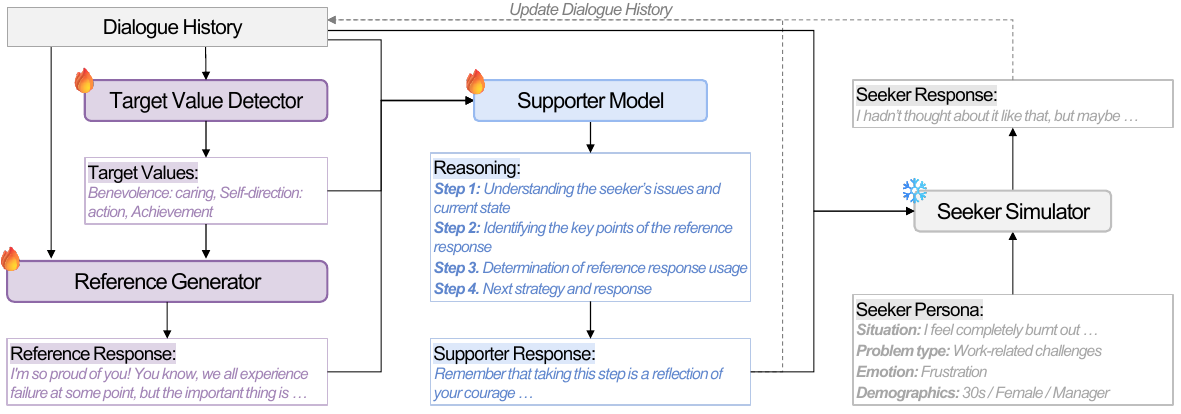}  
    \caption{Overview of the framework with three components: (1) \textbf{target value detector}, identifying values to reinforce in the seeker at each turn; (2) \textbf{reference generator}, producing reference responses to promote these values; and (3) \textbf{supporter model}, generating supporter's responses based on the target values and reference responses.}
    \label{fig:overall-framework} 
\end{figure*}

\section{Emotional Support Dataset from Reddit}
\label{sec:reddit-data}

Providing emotional support through value reinforcement involves addressing two critical questions: (1) which values should be reinforced at each turn, and (2) what supporter utterances can reinforce them most effectively. 
Addressing these questions requires large, authentic conversation data that span a wide range of help-seeking situations.
To that end, we turn to Reddit's \textit{r/offmychest} subreddit, which offers a diverse collection of emotional support exchanges. In this context, original posters (OPs) are seekers, and commenters serve as supporters. The structure of posts and comment threads closely mirrors dialogue flows, capturing the dynamics of emotional support interactions.
We collected posts and comments from 2019 to 2023, as provided by \citet{reddit-torrents}. 
We retained only high-quality emotional support conversations by filtering them using metrics such as upvote ratio and score. The collected data was limited to publicly available content and did not include private, deleted, or personally identifiable information.

Our goal is to use this data to train a model that identifies the values to reinforce at each turn (target value detector) and a model that produces supporter utterances to effectively promote the target values (reference generator).
For this purpose, we labeled the data with sentiment strength and expressed values at both the post and comment levels using models developed by \citet{liu2024emollms} and \citet{schroter-etal-2023-adam}.
Values expressed in a positive comment by the OP can be considered successful target values at that time, while the preceding comment from a commenter can be regarded as an effective supporter utterance that promotes those values.
The dataset contains over 20,000 samples, with details on the classification models and the generated dataset provided in Appendix~\ref{sec:appendix-reddit}.

\section{Method}
\label{sec:method}
The overall framework, illustrated in Figure~\ref{fig:overall-framework}, consists of three core components: (1) \textbf{target value detector} identifies values to reinforce at each turn; (2) \textbf{reference generator} produces utterances to effectively promote these values from the seeker; (3) \textbf{supporter model} determines strategies and generates responses based on the identified target values and the reference responses.

\subsection{Target Value Detector}

We train the target value detector using the emotional support conversations from Reddit (Section~\ref{sec:reddit-data}).
Given a dialogue history (\(o_1, c_1, o_2, c_2, ..., c_{t-1}, o_{t}\)), where \( o_i \) and \( c_i \) represent the $i$th utterances by the OP (seeker) and a commenter (supporter), respectively, the target value detector predicts which values to target in \(c_t\). 
The ground-truth values $v_{t+1}$ are the top-3 values observed in \(o_{t+1}\), based on their probabilities from the value detection model \citep{schroter-etal-2023-adam}.
\begin{equation}
v_{t+1} = \text{LM}_\text{TVD}(o_1, c_1, o_2, c_2, ..., c_{t-1}, o_t)
\end{equation}
Detailed training methods and results are provided in Appendix~\ref{sec:appendix-tvd}.

\subsection{Reference Generator}

The reference generator is also trained on the Reddit data. Specifically, given a dialogue history (\(o_1, c_1, o_2, c_2, ..., c_{t-1}, o_t\)) and the values (\(v_{t+1}\)) reflected in the OP's next utterance ($o_{t+1}$), the model is trained to generate \(c_t\).
Here, $v_{t+1}$ is treated as the target values and $c_t$ is considered to have successfully promoted these target values.
Training involves two stages: supervised fine-tuning (SFT) and direct preference optimization (DPO).

\paragraph{SFT Stage.} This stage involves training the model to generate the supporter's comments by conditioning on the dialogue history and the values expressed in the next utterance of the OP:
\begin{equation}
c_t = \text{LM}_\text{RG}(o_1, c_1, o_2, c_2, ..., c_{t-1}, o_t; v_{t+1})
\end{equation}

\paragraph{DPO Stage.} This stage aims to enhance the SFT model's generation quality through DPO.
The preference dataset is constructed as follows. 
Given a dialogue history (\(o_1, c_1, o_2, c_2, ..., c_{t-1}, o_t\)), the original supporter comment \( c_t \) is designated as the preferred response, as it successfully promoted the target values $v_{t+1}$. 
The rejected response is selected as another comment to $o_t$, denoted by $c_t'$, randomly sampled from the siblings of $c_t$ (i.e., other comments under the same dialogue history). $c_t'$ is a natural response to the dialogue history but is likely suboptimal for promoting the target values $v_{t+1}$ originally promoted by $c_t$.
To mitigate the risk that $c_t'$ is inadvertently effective for promoting $v_{t+1}$, we exclude any overlapping values between $v_{t+1}$ and $v_{t+1}'$ (i.e., the seeker's values expressed in $o_{t+1}'$ in response to $c_t'$), retaining up to three distinct target values unique to the preferred response in the final preference dataset.
Detailed training methods and results are provided in Appendix~\ref{sec:appendix-rg}.

\subsection{Supporter Model}
\label{sec:supporter}

The supporter model is the primary model that interacts with the seeker, generating responses that align with target values. It processes three key inputs: the dialogue history, the target values identified by the target value detector at each turn, and a reference response generated by the reference generator. At each turn, the model generates a response using the reasoning process across the following four aspects (Figure~\ref{fig:overall-framework}): (1) identifying the seeker’s issues and current state, (2) analyzing the key content of the reference response, (3) determining whether to incorporate the reference response into the final output, and (4) selecting the optimal emotional support strategy (Appendix~\ref{sec:es-strategies}) and generating the final response. The entire prompt is in Table~\ref{table:prompt-supporter}.
In step (3), the model generates either ``Yes'', along with an explanation of how the reference will be incorporated, or ``No'', with justification if the reference is deemed unsuitable.
This selective incorporation is necessary because, while Reddit data offers valuable information across diverse emotional support scenarios, its distribution may not always align with everyday conversations. We compare our method against the direct use of Reddit-based reference responses in Section~\ref{sec:exp-baselines}.

The training process involves two stages---SFT and DPO---using simulation data as follows.

\paragraph{SFT Stage.}
SFT requires large-scale emotional support conversations grounded in value reinforcement. To obtain such data, we opt to use dialogue simulation with a seeker simulator (Section~\ref{sec:seeker-simulator}). 
We use GPT-4o-mini for both the supporter and seeker simulators to generate data for training a smaller model.\footnote{In our pilot experiment, zero-shot Llama-3-8B-Instruct was found to be unsuitable as a supporter simulator due to issues like repetitive responses and biases in reference usage.}
The simulators engage in interactions by iteratively producing an utterance based on the ongoing dialogue history as a prompt and appending it to the history prompt. 

During simulations, GPT avoids using reference responses in approximately 90\% of cases. 
To prevent models fine-tuned on this data from inheriting the same bias, we simulate additional responses (called ``alternative responses'') at each supporter turn. 
Specifically, if GPT initially used the reference response, we simulate an alternative response without the reference response, and vice versa.

The simulated dialogues are employed to fine-tune Llama-3-8B-Instruct, with dataset sizes outlined in Table~\ref{table:sm-dataset-size}.

\begin{table}[t]
\centering
\resizebox{0.65\columnwidth}{!}{%
\begin{tabular}{@{}llrr@{}}
\toprule
\textbf{Stage} & \textbf{Supporter} & \textbf{Train} & \textbf{Dev} \\ \midrule
SFT & GPT-4o-mini & 33,130 & 2,367 \\
DPO & SFT & 3,301 & 628 \\
\bottomrule
\end{tabular}%
}
\caption{Dataset sizes for training the supporter model generated through simulation. The `Supporter' column refers to the supporter model used in the simulation.}
\label{table:sm-dataset-size}
\end{table}

\paragraph{\textbf{DPO Stage.}} 
We construct the preference data as follows.
For each dialogue, every supporter turn is assumed to have two response candidates (i.e., one with and one without using the reference response). 
We compute the expected reward for each response to determine the preferred and rejected responses for DPO. 
This reward is based on how many intended target values at that turn are expressed in the seeker's subsequent utterances.
The reward for a supporter response at turn $t$, $u_t^\text{sup}$, is:
\begin{equation}
R(u_t^\text{sup}) = \sum_{k=1}^{h} \gamma^{k-1} N_{t+k}
\label{equation:dpo-reward}
\end{equation}
where $N_{t+k}$ is the frequency of the values targeted at turn \(t\) appearing in the seeker's utterance at turn $t+k$, \(h\) is the look-ahead horizon (the number of future steps considered), and \( \gamma \) is a discount factor balancing immediate and future rewards. 
A response pair is added to the DPO dataset only if their reward difference exceeds a threshold \(T_\text{diff}\).

To prepare the dialogues underlying the preference data above, we conduct additional dialogue simulations between the SFT supporter model and the seeker simulator. This is because the SFT model has an enhanced ability to generate and explore diverse dialogue flows. Table~\ref{table:sm-dataset-size} summarizes the total dataset sizes, while hyperparameter details are presented in Table~\ref{table:sm-dpo-dataset}.
Details of the methods are provided in Appendix~\ref{sec:appendix-sm}. 

\begin{figure}[!t]
    \includegraphics[width=\columnwidth]{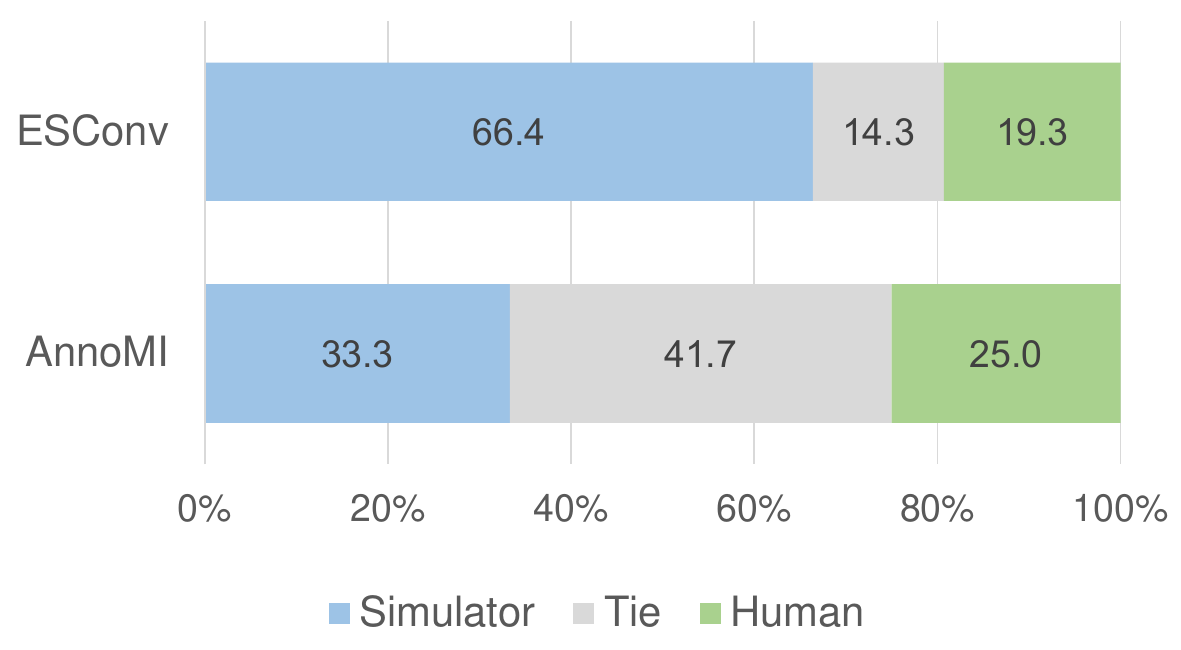}  
    \caption{Win ratios in human evaluation comparing the naturalness of responses from the seeker simulator and human seekers on the ESConv \citep{liu-etal-2021-towards} and AnnoMI \citep{annomi-2022, annomi-2023} datasets.}
    \label{fig:seeker-human-eval} 
\end{figure}

\subsection{Seeker Simulator}
\label{sec:seeker-simulator}

The seeker simulator generates seeker utterances based on the provided persona and dialogue history.
To simulate various scenarios, we generated personas using GPT-4o and GPT-4o-mini, defining attributes such as problem type, emotions, and situations (Figure~\ref{fig:overall-framework}), informed by prior studies \citep{liu-etal-2021-towards, zhao-etal-2024-esc}. The process resulted in 2,036 unique personas: 1,796 for training, 120 for development, and 120 for testing. 

The seeker simulator is based on GPT-4o-mini, with its detailed design and validity provided in Appendix~~\ref{sec:ref-seeker-simulator}. To summarize, we extensively verified the quality of the seeker simulator using human and automated evaluations. Human evaluators judged our seeker simulator to be as natural as, or more natural than, human seekers (Figure~\ref{fig:seeker-human-eval}). Additionally, utterances produced by GPT-4o-mini as the seeker simulator more closely resemble human seekers' utterances in content, emotional tone, and value alignment, compared to other models.

\section{Experiments}
We evaluate our model \textbf{ES-VR} (\textbf{E}motional \textbf{S}upport via \textbf{V}alue \textbf{R}einforcement) through comprehensive experiments.

\subsection{Evaluation Methods}
We evaluate various supporter models through conversations with the seeker simulator using the 120 held-out seeker personas for testing.
A conversation is considered complete if the seeker simulator generates ``[END]'' or if the seeker’s emotion score, as calculated by EmoLlama-Chat-7B \citep{liu2024emollms}, reaches 0.6 or higher with gratitude expressions (e.g., ``thank you''). Interactions are limited to a maximum of 20 turns, based on the average conversation length of 15 turns observed in the ESConv dataset. Only conversations concluding within this limit are included in the evaluation.

\subsection{Evaluation Metrics}

We conduct evaluations focusing on three key aspects: ES-Skills, ES-Intensity, and ES-Value. 
A detailed explanation of the metrics is in Appendix~\ref{sec:eval-metrics}.

\textbf{ES-Skills} evaluates a supporter's capabilities across three components, based on prior studies \citep{zheng-etal-2024-self, zhao-etal-2024-esc, cheng-etal-2023-pal, deng2024ppdpp, cheng-etal-2022-improving, liu-etal-2021-towards}: (1) emotional support skills, including \textit{Identification}, \textit{Comforting}, \textit{Suggestions}, \textit{Experience}, and \textit{Informativeness}; (2) general conversation skills, covering \textit{Consistency}, \textit{Role-Adherence}, \textit{Expression}, and \textit{Humanness}; and (3) an \textit{Overall}. Each criterion is rated on a five-point scale using GPT-4o-mini. 

\textbf{ES-Intensity} measures the intensity of a seeker's negative emotions after a conversation. Scores are assigned on a five-point scale, with lower scores indicating minimal negative emotions. We developed a predictive model using GPT-4o-mini based on ratings provided by human seekers in ESConv. The model demonstrates a correlation of 0.345 with the actual ratings. 

\textbf{ES-Value} assesses value reinforcement from two perspectives: the seeker's experience of value exploration and reinforcement within conversations, and the supporter's contribution to this process. We conduct pairwise comparisons between models using GPT-4o-mini as a judge. The reason is that, when assessing conversations individually on a 1--5 scale, GPT tends to award scores of 4 and 5 to most conversations, making it difficult to discern performance differences among models.

To validate our GPT-based evaluation for ES-Skills and ES-Value, we calculated correlations with ratings from licensed therapists. All criteria showed positive correlations (0.198--0.778), most of which were statistically significant (Appendix~\ref{sec:appendix-expert-corr}).

\begin{table*}[t]
\centering
\resizebox{\textwidth}{!}{%
\begin{tabular}{@{}l*{13}l@{}}
\toprule
\multirow{2.5}{*}{\makecell[c]{\textbf{Models}\\}} & \multicolumn{10}{c}{\textbf{ES-Skills}↑} & \multirow{2.5}{*}{\makecell[c]{\textbf{ES-}\\\textbf{Intensity}↓\\}} & \multicolumn{2}{c}{\textbf{ES-Value}$^{\clubsuit}$↑} \\ 
\cmidrule(lr){2-11} \cmidrule(lr){13-14}
 & \textbf{Iden.} & \textbf{Comf.} & \textbf{Sugg.} & \textbf{Expe.} & \textbf{Info.} & \textbf{Cons.} & \textbf{Role.} & \textbf{Expr.} & \textbf{Huma.} & \textbf{Over.} & & \textbf{Seeker} & \textbf{Supporter} \\ \midrule
GPT-4o-mini               & \hspace{1mm}4.77 & \hspace{1mm}4.88 & \hspace{1mm}4.03\textsuperscript{*} & \hspace{1mm}2.34$^{*}$ & \hspace{1mm}4.11$^{*}$ & \hspace{1mm}4.98 & \hspace{1mm}\textbf{5.00} & \hspace{1mm}3.97$^{*}$ & \hspace{1mm}4.45$^{*}$ & \hspace{1mm}4.44$^{*}$ & 
\hspace{4mm}2.19$^{*}$ &
\hspace{2mm}0.43$^{*}$ & \hspace{4mm}0.36$^{*}$ 
\\
+ Target values           & \hspace{1mm}4.83 & \hspace{1mm}4.88 & \hspace{1mm}4.38$^{*}$ & \hspace{1mm}2.48$^{*}$ & \hspace{1mm}4.27$^{*}$ & \hspace{1mm}4.99 & \hspace{1mm}\textbf{5.00} & \hspace{1mm}4.01$^{*}$ & \hspace{1mm}4.53$^{*}$ & \hspace{1mm}4.59$^{*}$ & 
\hspace{4mm}1.96 &
\hspace{2mm}0.48 & \hspace{4mm}0.48
\\
+ Reference               & \hspace{1mm}4.82 & \hspace{1mm}4.91 & \hspace{1mm}4.34$^{*}$ & \hspace{1mm}2.54$^{*}$ & \hspace{1mm}4.29$^{*}$ & \hspace{1mm}\textbf{5.00} & \hspace{1mm}\textbf{5.00} & \hspace{1mm}4.02$^{*}$ & \hspace{1mm}4.55$^{*}$ & \hspace{1mm}4.61$^{*}$ & 
\hspace{4mm}\textbf{1.89} &
\hspace{2mm}0.47$^{*}$ & \hspace{4mm}0.42$^{*}$
\\
+ Both & \hspace{1mm}\textbf{4.83} & \hspace{1mm}\textbf{4.92} & \hspace{1mm}\textbf{4.57} & \hspace{1mm}\textbf{3.11} & \hspace{1mm}\textbf{4.42} & \hspace{1mm}\textbf{5.00} & \hspace{1mm}\textbf{5.00} & \hspace{1mm}\textbf{4.10} & \hspace{1mm}\textbf{4.70} & \hspace{1mm}\textbf{4.72} & 
\hspace{4mm}\textbf{1.89} &
\hspace{5mm}- & \hspace{7mm}- 
\\ \bottomrule
\end{tabular}%
}
\caption{Emotional support performance depending on the incorporation of target value information and reference responses. $^{\clubsuit}$\textbf{ES-Value}: The win-ratio of each model against \textit{GPT-4o-mini (Both)}; a value lower than 0.5 means the model lost more often than it won against \textit{GPT-4o-mini (Both)}. Statistically significant differences compared to \textit{GPT-4o-mini (Both)} are indicated with * ($p$-value $< 0.05$) based on the Mann-Whitney U test.}
\label{table:gpt-ablation-result1}
\end{table*}

\subsection{Baselines}

\begin{itemize}[topsep=5pt, partopsep=0pt, itemsep=3pt, parsep=0pt] 
    \item \textbf{Prompt-Based}: GPT-4o-mini and Llama-3-8B-Instruct. 
    \item \textbf{ES Datasets}: Variants of Llama-3-8B-Instruct trained on emotional support datasets, including Reddit (Section~\ref{sec:reddit-data}), ESConv \citep{liu-etal-2021-towards}, ExTES \citep{zheng-etal-2024-self}, and Psych8k \citep{liu2023chatcounselor}. 
    \item \textbf{ES Methods}: Recent methods contributing to the development of supporter models, including Ask-an-Expert \citep{zhang-etal-2023-ask}, ESCoT \citep{zhang-etal-2024-escot}, and PPDPP \citep{deng2024ppdpp}. Details of these methods are provided in Appendix~\ref{sec:es-method-driven}.
    \item \textbf{Emotion-Reinforced}: To verify the effectiveness of value reinforcement, we train the reference generator and supporter model to reinforce positive emotions instead of values (Appendix~\ref{sec:emo-reinforced}).
\end{itemize}

\subsection{Evaluation Results}

\subsubsection{Effectiveness of Value Targeting and Reference Responses}
To evaluate the impact of our two main components, target value prediction and reference response generation, we first conducted an ablation study using GPT-4o-mini as the supporter model. 

As shown in Table~\ref{table:gpt-ablation-result1}, leveraging both target values and reference responses significantly improved performance across all ES-Skills metrics while reducing ES-Intensity. This approach notably enhanced key ES-Skills, including \textit{Suggesting}, \textit{Expression}, and \textit{Informativeness}. Similarly, value reinforcement was substantially improved when both target values and reference responses were utilized. These findings emphasize the effectiveness of targeting specific values at each turn and using reference responses that leverage real-world knowledge from Reddit.
Since our fine-tuned models are trained on GPT-simulated data, we use the simulation data that incorporates both target values and reference responses in subsequent experiments.

\begin{table*}[ht]
\centering
\resizebox{\textwidth}{!}{%
\begin{tabular}{@{}ll*{12}{l}l@{}}
\toprule
\multirow{2.5}{*}{\makecell[c]{\textbf{Categories}}} & \multirow{2.5}{*}{\makecell[c]{\textbf{Models}}} & \multicolumn{10}{c}{\textbf{ES-Skills}↑} & 
\multirow{2.5}{*}{\makecell[c]{\textbf{ES-}\\\textbf{Intensity}↓}} &
\multicolumn{2}{c}{\textbf{ES-Value}$^{\clubsuit}$↑} \\ 
\cmidrule(lr){3-12} \cmidrule(lr){14-15} 
& & \textbf{Iden.} & \textbf{Comf.} & \textbf{Sugg.} & \textbf{Expe.} & \textbf{Info.} & \textbf{Cons.} & \textbf{Role.} & \textbf{Expr.} & \textbf{Huma.} & \textbf{Over.} & & \textbf{Seeker} & \textbf{Supporter}\\ \midrule

\multirow{1}{*}{\makecell[l]{{Prompt-}\\{Based}}} &
GPT & 4.83$^{*}$ & 4.92 & 4.57$^{*}$ & 3.11$^{*}$ & 4.42$^{*}$ & \textbf{5.00} & \textbf{5.00} & 4.10$^{*}$ & 4.70$^{*}$ & 4.72$^{*}$ & 
\hspace{4mm}1.89$^{*}$ &
\hspace{2mm}0.49$^{*}$ & \hspace{4mm}0.42$^{*}$ 
\\
& Llama & 4.87 & 4.91 & 4.43$^{*}$ & 2.91$^{*}$ & 4.47$^{*}$ & 4.99 & \textbf{5.00} & 4.03$^{*}$ & 4.63$^{*}$ & 4.68$^{*}$ & 
\hspace{4mm}1.99$^{*}$ &
\hspace{2mm}0.46$^{\dagger}$ & \hspace{4mm}0.45 
\\ \midrule

\multirow{1}{*}{\makecell[l]{{ES Datasets}}} &
Llama-Reddit & 3.38$^{*}$ & 3.74$^{*}$ & 3.21$^{*}$ & 2.59$^{*}$ & 2.99$^{*}$ & 3.94$^{*}$ & 4.35$^{*}$ & 3.37$^{*}$ & 3.81$^{*}$ & 3.40$^{*}$ & 
\hspace{4mm}1.97$^{*}$ &
\hspace{2mm}0.29$^{*}$ & \hspace{4mm}0.09$^{*}$ 
\\
& Llama-ESConv & 4.35$^{*}$ & 4.43$^{*}$ & 4.06$^{*}$ & 2.65$^{*}$ & 3.88$^{*}$ & 4.82$^{*}$ & 4.97$^{*}$ & 3.79$^{*}$ & 4.25$^{*}$ & 4.22$^{*}$ & 
\hspace{4mm}1.87$^{\dagger}$ &
\hspace{2mm}0.37$^{*}$ & \hspace{4mm}0.19$^{*}$ 
\\
& Llama-ExTES & 4.83$^{*}$ & 4.90$^{\dagger}$ & 4.53$^{*}$ & 2.71$^{*}$ & 4.44$^{*}$ & 4.99 & \textbf{5.00} & 4.02$^{*}$ & 4.59$^{*}$ & 4.66$^{*}$ & 
\hspace{4mm}\underline{1.67} &
\hspace{2mm}0.48$^{\dagger}$ & \hspace{4mm}0.51
\\
& Llama-Psych8k & 4.84$^{*}$ & 4.85$^{*}$ & 4.75$^{*}$ & 2.89$^{*}$ & \underline{4.63}$^{\dagger}$ & 4.99 & \textbf{5.00} & 4.05$^{*}$ & 4.57$^{*}$ & 4.75$^{*}$ & 
\hspace{4mm}\textbf{1.53}$^{*}$ &
\hspace{2mm}0.49 & \hspace{4mm}0.62$^{*}$ 
\\ \midrule

\multirow{1}{*}{\makecell[l]{{ES Methods}}} 
& Ask-an-Expert & 4.13$^{*}$ & 4.30$^{*}$ & 3.93$^{*}$ & 3.12$^{*}$ & 3.70$^{*}$ & 4.61$^{*}$ & 4.91$^{*}$ & 3.74$^{*}$ & 4.21$^{*}$ & 4.08$^{*}$ & 
\hspace{4mm}1.86 &
\hspace{2mm}0.32$^{*}$ & \hspace{4mm}0.15$^{*}$ 
\\
& ESCoT & 3.69$^{*}$ & 3.91$^{*}$ & 3.16$^{*}$ & 1.81$^{*}$ & 3.07$^{*}$ & 4.16$^{*}$ & 4.81$^{*}$ & 2.95$^{*}$ & 3.64$^{*}$ & 3.51$^{*}$ & 
\hspace{4mm}2.25$^{*}$ &
\hspace{2mm}0.25$^{*}$ & \hspace{4mm}0.05$^{*}$ 
\\
& PPDPP & 4.64$^{*}$ & 4.88$^{*}$ & 4.45$^{*}$ & 2.49$^{*}$ & 4.26$^{*}$ & 4.99 & \textbf{5.00} & 3.99$^{*}$ & 4.54$^{*}$ & 4.54$^{*}$ & 
\hspace{4mm}1.83 &
\hspace{2mm}0.44$^{*}$ & \hspace{4mm}0.31$^{*}$ 
\\
\midrule

\multirow{1}{*}{\makecell[l]{{Emotion-}\\{Reinforced}}} &
SFT & 4.83$^{\dagger}$ & 4.91 & 4.51$^{*}$ & 3.64$^{\dagger}$ & 4.43$^{*}$ & 4.97$^{*}$ & 4.99 & 4.16$^{*}$ & 4.67$^{*}$ & 4.73$^{*}$ & 
\hspace{4mm}1.97$^{*}$ &
\hspace{2mm}0.49 & \hspace{4mm}0.46$^{\dagger}$ 
\\
& DPO & 4.85 & \underline{4.92} & 4.74 & \textbf{4.05}$^{\dagger}$ & 4.61 & 4.99 & \textbf{5.00} & \textbf{4.33} & \textbf{4.78} & 4.82 & 
\hspace{4mm}1.86 &
\hspace{2mm}0.49 & \hspace{4mm}0.51 
\\ \midrule

\multirow{1}{*}{\makecell[l]{{ES-VR}\\{(Ours)}}} &
SFT & 4.85$^{\dagger}$ & 4.90 & 4.72$^{*}$ & 3.76 & 4.56$^{*}$ & 4.99 & \textbf{5.00} & 4.25 & 4.73 & 4.80$^{*}$ & 
\hspace{4mm}1.86 &
\hspace{2mm}0.48$^{\dagger}$ & \hspace{4mm}0.46$^{\dagger}$ 
\\

& DPO & \textbf{4.90} & \textbf{4.95} & \textbf{4.80} & \underline{3.85} & \textbf{4.69} & \textbf{5.00} & \textbf{5.00} & \underline{4.30} & \underline{4.77} & \textbf{4.87} & 
\hspace{4mm}1.75 &
\hspace{5mm}- & \hspace{7mm}- 
\\ 
\cmidrule(lr){2-15}

& DPO (Cactus) & \underline{4.89} & 4.90 & \underline{4.76} & 2.72 & 4.60 & 4.99 & \textbf{5.00} & 4.03 & 4.60 & \textbf{4.87} & \hspace{4mm}1.75 &
\hspace{5mm}- & \hspace{7mm}- 
\\ 
\bottomrule

\end{tabular}%
}
\caption{Comparison of models based on ES-Skills and ES-Intensity. $^{\clubsuit}$\textbf{ES-Value}: The win-ratio of each model against \textit{ES-VR (DPO)}. Statistically significant differences compared to our DPO model are marked with * ($p$-value $< 0.05$), and differences with $p$-value $< 0.1$ are marked with $\dagger$, as determined by the Mann-Whitney U test.}
\label{table:exp2-baseline-comparison}
\end{table*}

\subsubsection{Performance Comparison with Baselines}
\label{sec:exp-baselines}
The performance comparisons between our models and the baselines are presented in Table~\ref{table:exp2-baseline-comparison}. For our DPO and emotion-reinforced DPO models, we select optimal configurations (\(h=3\), \( \gamma=1 \),  \(T_\text{diff}=2\) and \(h=3\), \( \gamma=1 \),  \(T_\text{diff}=0.5\), respectively). For results with more DPO hyperparameters, refer to Table~\ref{table:exp2-baseline-comparison-all} in the Appendix.

\paragraph{ES-Skills.} Our DPO model outperformed the baselines across most metrics, particularly in emotional support metrics such as \textit{Suggestions}, \textit{Experience}, and \textit{Informativeness}. These improvements reflect the characteristics of our reference responses, which emphasize sharing relevant experiences and offering practical solutions---key elements of effective online emotional support. The models also demonstrated significant gains in conversational capabilities, especially in \textit{Expression} and \textit{Humanness}, resulting in more natural and engaging interactions.

Notably, the variant of our method that focuses on reinforcing positive emotions rather than values (\textit{Emotion-Reinforced}) also consistently outperformed other baselines. This suggests that one of our key ideas---leveraging crowd knowledge from Reddit---is still effective when the supporter model is designed to promote positive emotions in seekers. Yet, our value reinforcement approach achieved higher scores across most emotional support skill metrics at comparable training stages, highlighting the effectiveness of reinforcing values in enhancing emotional support.

\paragraph{ES-Intensity.} Our DPO model outperformed most baselines, demonstrating that our supporter model reduces the intensity of seekers' negative emotions more effectively than other methods. 
Notably, our models achieved lower intensity levels than the emotion reinforcement models at comparable training stages.
This suggests that redirecting a seeker’s focus to values can indirectly alleviate negative emotions, highlighting a promising direction for future research.

Llama-Psych8k showed significantly lower ES-Intensity than our model. Analysis revealed it generated much longer responses (73 words per turn) compared to other models (20–25 words). 
Since the ES-Intensity model was validated on ESConv, caution is warranted when interpreting the scores of dialogues with substantially different distributions. 
Moreover, in practice, its lengthy responses and lower \textit{Humanness} scores may feel overwhelming, discouraging seeker engagement.

\paragraph{ES-Value.} Table~\ref{table:exp2-baseline-comparison} presents the win ratios of baselines against our DPO model for ES-Value (detailed results are provided in Table~\ref{table:exp2-es-value-all}). Our models outperformed the baselines in most comparisons, highlighting their effectiveness of eliciting seekers' values.
Exception was observed in evaluations from the supporter’s perspective against Llama-Psych8k. Upon review, this result seems to be attributed to its long responses, which include a large amount of content potentially related to values. Expert therapists determined that GPT tends to evaluate this model more favorably than other models. 
Moreover, our model exhibited slightly better performance from the seeker’s perspective. 

Despite the strong performance of our DPO model compared to most baselines, its results were comparable to those of the emotion-reinforcing DPO and certain fine-tuned models. An analysis of 40 dialogues with low ES-Value scores revealed two key areas for improvement: enhancing the ability to identify seekers’ unique strengths and accomplishments, and improving the capacity to address their emotional states and concerns more deeply.
These findings underscore the need to improve the model's engagement with seekers’ individual circumstances, which is expected to enhance its value reinforcement performance. Detailed experimental methods and results are provided in Appendix~\ref{sec:appendix-indepth-value}.

\begin{table}[t]
\centering
\resizebox{0.8\columnwidth}{!}{%
\begin{tabular}{@{}ll*{3}{c}@{}}
\toprule
\multirow{2.5}{*}{\makecell[c]{\textbf{Categories}}} & \multirow{2.5}{*}{\makecell[c]{\textbf{Models}}} & 
\multicolumn{3}{c}{\textbf{Valid Turns}} \\
\cmidrule(lr){3-5} 
& & \textbf{1} & \textbf{2} & \textbf{3} \\ \midrule

\multirow{1}{*}{\makecell{{Prompt-Based}}} &
GPT & 0.681 & 0.728 & 0.740 \\
& Llama & 0.688	& 0.728 & 0.751 \\
\midrule

\multirow{1}{*}{\makecell{{ES-VR (Ours)}}} &
DPO & \textbf{0.720} & \textbf{0.760} & \textbf{0.779} \\
\bottomrule

\end{tabular}%
}
\caption{Average success rates of target value reinforcement within 1, 2, and 3 valid turns---the number of future turns within which the target values remained relevant.}
\label{table:value-reinforcement-results}
\end{table}

\subsubsection{Success of Target Value Reinforcement}

As our approach focuses on setting target values and reinforcing them, we evaluated its effectiveness in reinforcing the target values. To this end, we compared our ES-VR (DPO) model with other prompt-based models that also take target values and reference responses as inputs.

Table~\ref{table:value-reinforcement-results} presents the average success rates across models, showing that our model consistently outperforms others in all cases for the next 1--3 turns.
Since all models use the same inputs---target values and a reference response---these differences underscore the effectiveness of our supporter model’s training methods in reinforcing target values.
Further details of this analysis are provided in Appendix~\ref{sec:value-reinforcement}.


\subsubsection{Generalization beyond Reddit Data}

To evaluate the generalizability of our target value detection and reference generation methods beyond Reddit, we explored training these two models on the Cactus dataset \citep{lee-etal-2024-cactus}---counseling conversations based on \textit{Cognitive Behavioral Theory (CBT)}---instead of Reddit. Detailed procedures are provided in Appendix~\ref{sec:generalization}.

For ES-Value (shown in Table~\ref{table:cactus-es-value} for clarity), DPO (Cactus) outperformed all baselines, demonstrating the supporter model’s ability to effectively learn and apply value signals from Cactus. This result confirms the generalizability of our target value detection and reference response generation approach.

Similarly, for ES-Skills (Table~\ref{table:exp2-baseline-comparison}), DPO (Cactus) performed comparably to or better than the baselines, particularly excelling in emotional support metrics. However, it scored significantly lower on the Experience metric compared to the original DPO model based on Reddit. Since Reddit contains numerous shared personal narratives, the reference response generator appears to benefit more from Reddit than from Cactus, supporting our decision to use Reddit.

\subsubsection{Expert Evaluation}
\label{sec:expert-eval}

To gain deeper insights into the value reinforcement capabilities of our supporter models, two licensed clinical psychologists with over three years of clinical experience conducted a qualitative analysis of dialogues generated by our ES-VR (DPO) model.

\paragraph{Strengths.} One notable strength is its ability to effectively validate the seeker's challenges, using empathetic phrases such as \textit{``which is completely understandable''}. This validation fosters trust between the supporter and the seeker while encouraging self-acceptance, which in turn promotes deeper exploration and understanding of personal values.

Another strength is our model's capacity to emphasize positive aspects of the seeker's situation, reflecting positive values and related goal. For example, responses like \textit{``Your initiative to seek meaningful experiences reflects your dedication to making a difference, and that determination will surely lead you closer to your goals.''} help seekers recognize their strengths and positive attributes. This behavior was contrasted with GPT-4o's responses, particularly when the seeker persisted in a negative mood. GPT-4o tended to focus heavily on expressing empathy and lingered in the negative mood. This overemphasis on empathy is likely a result of the human preference alignment process. Our method overcomes this tendency by enhancing the seeker's self-awareness and supporting the reinforcement of their values in a constructive manner.

\paragraph{Areas for Improvement.} To enhance the effectiveness of value reinforcement, three key improvements are recommended. First, deeper understanding of seekers' perspectives and circumstances would allow for more tailored support. Second, addressing potential obstacles associated with pursuing values would help equip seekers to navigate practical challenges. Finally, offering clear definitions and concrete examples of proposed values, while encouraging seekers to articulate their own interpretations, would strengthen the connection between abstract values and lived experiences.

\section{Conclusion}

In this paper, we introduce the first emotional support framework based on value reinforcement, as emphasized in modern psychotherapy. The framework incorporates a target value detector and a reference generator to improve the supporter model’s ability to generate value-aligned and effective support responses. Evaluations demonstrate that our framework surpasses baseline models in both emotional support quality and value reinforcement. Expert therapist evaluations further highlight the model’s strengths in validating seekers’ challenges and emphasizing positive aspects of their situations, which are key elements of effective emotional support. These results underscore the potential of value reinforcement to enhance supportive interactions and provide a foundation for developing more effective emotional support systems.

\section*{Limitations}
Our framework demonstrates promising results in enhancing emotional support quality and reinforcing values. However, there is a limitation in the lack of longitudinal evaluation. While previous research highlights the long-term benefits of value reinforcement in counseling and decision-making, the long-term outcomes of our framework have yet to be empirically validated. Future studies could incorporate extended timeframes to evaluate its sustained impact on emotional well-being and guide further refinements.

Our model demonstrated superior performance in value reinforcement evaluations, outperforming most baselines in pairwise comparisons. However, Section~\ref{sec:expert-eval} highlights some areas for improvement, particularly in the deep exploration of seekers' issues and thoughts, as well as in addressing potential obstacles and setbacks. Future research should prioritize these aspects by developing more comprehensive datasets and advancing training methodologies.

In this study, simulations for DPO training of the supporter model focused on varying conversation paths based solely on the use of reference responses, with rewards evaluated in terms of value reinforcement. However, other factors, such as strategy selection, may also significantly impact value reinforcement. We anticipate that incorporating these additional factors into future simulations and training could further enhance the performance of the supporter model.

\section*{Ethical Considerations}

\subsection*{Considerations on Self-Disclosure}

Sharing experiences related to those of the seeker is a key strategy in emotional support for fostering intimacy and has been a key evaluation criterion in prior emotional support systems \citep{zhang-etal-2023-ask}. However, some users might feel uncomfortable when dialogue systems present these experiences as personal. We found that removing self-disclosure strategy from the model impacts the quality of emotional support (Appendix~\ref{sec:appendix-self-disclosure}), highlighting the need for further research into more sophisticated approaches to experience sharing, which we leave as a direction for future work.

\subsection*{Potential Risks of Misuse or Harm}
Our system provides emotional support for common daily challenges, such as interpersonal conflicts and academic stress, while explicitly not replacing professional psychological intervention. Although automated and expert evaluations demonstrate strong performance, there is a possibility that the system’s responses might inadvertently have an unintended impact on users in certain situations. To mitigate this risk, we have implemented mechanisms for context-sensitive responses and clearly positioned the system as a supplementary tool rather than a substitute for professional therapy.

\subsection*{Addressing Bias and Overgeneralization}
Data from online platforms inherently contains biases that may underrepresent certain perspectives, potentially limiting the system's ability to effectively serve diverse user groups. To address these concerns, we carefully selected data collection targets and periods to ensure diversity in emotional support topics. Additionally, we enabled the supporter model to evaluate the appropriateness of reference responses, introducing an additional filtering process. By fostering balanced viewpoints, we aim to provide equitable and inclusive support.

\section*{Acknowledgments}
This work was supported by the New Faculty Startup Fund and the Creative-Pioneering Researchers Program from Seoul National University. It was also supported by the National Research Foundation of Korea (NRF) grants funded by the Korean government (MSIT) (RS-2024-00333484, RS-2024-00455484).

\bibliography{custom}

\appendix

\section{Experiment Details for Identifying the Effect of Values in Emotional Support}
\label{sec:appendix-value-effect}

In this experiment, we compare positive value expressions in the seeker’s final four turns between \textit{high effectiveness} and \textit{low effectiveness groups}.
This analysis window is chosen based on three key considerations: (1) To control for variability in dialogue length within ESConv, we analyze a fixed number of turns. (2) To capture changes resulting from emotional support conversations, we focus on the latter part of the dialogue. (3) Given that the minimum number of turns in ESConv is eight, we set the analysis window at half this value.

Since we are interested in cases where values are expressed positively, we count the values only when the seeker’s utterances are positive. To determine whether an utterance is positive, we use the EmoLlama-Chat-7B proposed by \citet{liu2024emollms}. To extract the values in the seeker’s utterances, we employ the model that achieved the best performance in the \textit{SemEval2023 Task 4: Identification of Human Values behind Arguments} \citep{schroter-etal-2023-adam, kiesel-etal-2023-semeval}.

\section{Constructing the Emotional Support Dataset from Reddit}
\label{sec:appendix-reddit}

\subsection{Models for Sentiment Strength and Values Labeling}

Automated classification models are essential for labeling Reddit data at scale. 
We prioritize models with optimal performance, as classification accuracy directly influences the quality of subsequent fine-tuning.

\paragraph{Sentiment Strength.} We employ EmoLlama-Chat-7B \citep{liu2024emollms}, which demonstrates superior performance in the V-reg task, estimating emotional valence on a continuous scale from 0 (most negative) to 1 (most positive). The comparative performance of various models, including both fine-tuned models and zero/few-shot methods, is presented in Table~\ref{table:emollama-performance}.

\paragraph{Human Values.} We adopt the top-performing model \citep{schroter-etal-2023-adam} from \textit{SemEval 2023 Task 4: Identification of Human Values behind Arguments} \citep{kiesel-etal-2023-semeval}, a shared task highly relevant to our objective based on the same value categorization. Considering recent advancements in LLMs, we also compare its performance against GPT-4o-mini by experimenting with two strategies: (1) predicting the binary presence of each value for a given sentence, and (2) assigning a probability score (between 0 and 1) to indicate the level of support for each value. 
As shown in Table~\ref{table:value-model-performance}, evaluation on the shared task’s test dataset demonstrates that the model from the shared task consistently outperforms GPT-4o-mini, supporting our decision to adopt it as our classification model. For reference, the F1-score of random classification is 0.128.

\begin{table}[t]
    \centering
    \resizebox{0.9\columnwidth}{!}{
        \begin{tabular}{l l c}
            \toprule
            \textbf{Category} & \textbf{Model} & \textbf{Corr.} \\
            \hline
            \multirow{3}{*}{Fine-tuned} 
                & BERT-base        & 0.840 \\
                & RoBERTa-base     & 0.845 \\
                & SentiBERT        & 0.835 \\
            \hline
            \multirow{9}{*}{\makecell[l]{{Zero-shot/}\\{Few-shot}}}
                & Falcon           & 0.135 \\
                & Vicuna           & 0.298 \\
                & LLaMA2-7B-chat   & 0.094 \\
                & LLaMA2-13B-chat  & 0.312 \\
                & ChatGPT          & 0.637 \\
                & ChatGPT-FS       & 0.739 \\
                & GPT-4            & 0.811 \\
                & GPT-4-FS         & 0.825 \\
            \hline
            Instruction-tuned & EmoLlama-Chat-7B & \textbf{0.876} \\
            \bottomrule
        \end{tabular}
    }
    \caption{Performance comparison of models on the V-reg task, measured using Spearman correlation coefficients. Results are from \citet{liu2024emollms}}
    \label{table:emollama-performance}
\end{table}

\begin{table}[t]
\centering
\resizebox{0.7\columnwidth}{!}{
\begin{tabular}{@{}lc@{}}
\toprule
\textbf{Models} & \multicolumn{1}{c}{\textbf{F1-Score}} \\ \midrule
Model from Shared Task & 0.57 \\
GPT (Binary) & 0.38 \\
GPT (Prob.) & 0.40 \\ \bottomrule
\end{tabular}
}
\caption{Performance comparison of the best-performing shared task model and GPT-4o-mini in a multi-label classification task across 20 values.}
\label{table:value-model-performance}
\end{table}

\subsection{Dataset Settings in Reddit}

The final datasets derived from Reddit are categorized into two settings:

\begin{itemize}[topsep=5pt, partopsep=0pt, itemsep=3pt, parsep=0pt]
    \item \textbf{Single-turn setting:} A concise, three-part interaction sequence consisting of an initial post (\textit{OP}), a response (\textit{commenter}), and a final reply (\textit{OP}).
    \item \textbf{Multi-turn setting:} Extended dialogue threads that include additional exchanges beyond the single-turn structure.
\end{itemize}

\noindent An overview of the final dataset is presented in Table~\ref{tab:reddit-data-dist}.

\section{Strategies for Emotional Support}
\label{sec:es-strategies}

In this study, we utilized the 8 emotional support strategies defined by \citet{liu-etal-2021-towards}. Descriptions of each strategy are as follows:

\begin{itemize}[topsep=5pt, partopsep=0pt, itemsep=3pt, parsep=0pt]
    \item \textbf{Question:} Ask open-ended or specific questions to help the seeker articulate and clarify the issues they are facing.
    \item \textbf{Restatement:} Rephrase the seeker’s statements in a clearer, more concise way to help them better understand their situation.
    \item \textbf{Reflection:} Express and describe the emotions that the seeker is experiencing to validate their feelings.
    \item \textbf{Self-disclosure:} Share similar experiences or emotions to convey empathy and build connection with the seeker.
    \item \textbf{Affirmation:} Highlight the seeker’s strengths and abilities while offering encouragement and reassurance.
    \item \textbf{Suggestions:} Offer practical advice or actionable steps to the seeker.
    \item \textbf{Information:} Share useful facts, resources, or data to help the seeker make informed decisions or gain clarity.
    \item \textbf{Others:} Use other support strategies that do not fall into the above categories.
\end{itemize}

\begin{table}[t]
\centering
\resizebox{0.8\columnwidth}{!}{
\begin{tabular}{@{}lrrrr@{}}
\toprule
\textbf{} & \multicolumn{1}{c}{\textbf{Train}} & \multicolumn{1}{c}{\textbf{Dev}} & \multicolumn{1}{c}{\textbf{Test}} & \multicolumn{1}{c}{\textbf{Total}} \\ \midrule
\textbf{Single-turn} & 18,459 & 2,000 & 1,000 & 21,459 \\
\textbf{Multi-turn} & 24,339 & 2,000 & 1,000 & 27,339 \\ \bottomrule
\end{tabular}
}
\caption{Data distribution of single-turn and multi-turn threads sourced from Reddit.}
\label{tab:reddit-data-dist}
\end{table}

\section{Training Details and Results}

\subsection{Target Value Detector}
\label{sec:appendix-tvd}

\paragraph{Training Methods.} The target value detector predicts the values observable in the next turn of the OP's comment. The model generates a sequence of values, and the top three are selected based on their predicted probabilities when multiple values are identified (\textit{e.g., "Self-direction: action, Benevolence: caring, Security: personal"}). 

The target value detector is based on the Llama-3-8B-Instruct, fine-tuned using the LoRA \citep{hu2022lora}. During training, the low-rank matrix dimension was set to 8, with an alpha of 16, and a learning rate of 5e-5. The final model was selected based on the highest F1-score achieved on the test dataset. Training was performed on an NVIDIA A100-80GB GPU, with durations of approximately 10 hours. The detailed training prompt is provided in Table~\ref{table:prompt-tvd}. 

\paragraph{Results.} The results of the training are summarized in Table~\ref{table:results-tvd}, comparing the performance of the target value detector with baseline models, GPT-4o-mini and Llama-3-8B-Instruct (vanilla). For the baselines, additional experiments were conducted by incorporating reasoning steps before response generation or providing detailed definitions for each value. The target value detector outperformed the baselines across all three metrics, demonstrating impressive performance considering the large set of 20 values.

\begin{table}[t]
\centering
\resizebox{\columnwidth}{!}{%
\begin{tabular}{@{}lccc@{}}
\toprule
\textbf{Models} & \textbf{Precision} & \textbf{Recall} & \textbf{F1-score} \\ \midrule
GPT-4o-mini              & 0.361 & 0.384 & 0.372 \\
\ + Reasoning            & 0.320 & 0.339 & 0.329 \\
\ + Value information    & \underline{0.383} & \underline{0.407} & \underline{0.395} \\ \midrule
Llama-3-8B-Instruct      & 0.323 & 0.283 & 0.302 \\
\ + Reasoning            & 0.304 & 0.283 & 0.293 \\
\ + Value information    & 0.343 & 0.271 & 0.303 \\ \midrule
Target Value Generator   & \textbf{0.516} & \textbf{0.540} & \textbf{0.528} \\ \bottomrule
\end{tabular}%
}
\caption{Performance comparison of models in target value prediction.}
\label{table:results-tvd}
\end{table}

\subsection{Reference Generator}
\label{sec:appendix-rg}

\paragraph{Training Methods.} The reference generator is based on Llama-3-8B-Instruct. The reference response model is based on Llama-3-8B-Instruct. Training was conducted on both single-turn and multi-turn settings using the Reddit dataset introduced in Section~\ref{sec:reddit-data}. For each setting, both SFT and DPO approaches were applied with various hyperparameter configurations. The model was trained for up to 5 epochs, and the final model was selected based on its performance on the test dataset. The hyperparameters used for the final model are summarized in Table~\ref{table:hyperparams-rg}. Training was performed on an NVIDIA A6000-48GB GPU, with durations of approximately 20 hours for the SFT stage and 10 hours for the DPO stage. Detailed training prompts are provided in Table~\ref{table:prompt-rg}.

\begin{table}[t]
\centering
\resizebox{\columnwidth}{!}{%
\begin{tabular}{@{}lccccc@{}}
\toprule
\textbf{Settings} & \textbf{Stage} & \textbf{LR} & \textbf{Rank}  & \textbf{Alpha}  & \textbf{Dropout} \\ \midrule
Single-turn & SFT & 1e-4 & 8 & 8 & 0.1 \\
& DPO & 1e-5 & 8 & 16 & 0.05 \\
\midrule
Multi-turn & SFT & 1e-5 & 8 & 16 & 0.05 \\
& DPO & 1e-5 & 8 & 16 & 0.05 \\
\bottomrule
\end{tabular}%
}
\caption{Hyperparameters used for training the reference generator.}
\label{table:hyperparams-rg}
\end{table}

\paragraph{Results.} The model performances were evaluated using GPT (GPT-4o-mini) through two approaches. First, pairwise comparisons were conducted between ``Llama-3-8B-Instruct (vanilla)-reference generator (SFT)'' and ``reference generator (SFT)-reference generator (DPO)''. Specifically, GPT assessed which model's responses more closely aligned with the ground truth responses (i.e., actual comments written by the original commenter) for the test dataset. Second, the impact of target values on the generated responses was examined. Responses generated using the original target values were compared to those generated using randomly assigned values to evaluate variation in content. To reduce sequence-based bias, the order of options within the prompts was alternated during evaluation. The evaluation prompts are detailed in Table~\ref{table:prompt-rg-evaluation}, and the results of the two experiments are presented in Table~\ref{table:results-rg-exp1} and Table~\ref{table:results-rg-exp2}, respectively.

The results indicate that in single-turn settings, the reference generator performed effectively in both experiments. In the first experiment, the reference generator (SFT) outperformed the baseline, while the reference generator (DPO) demonstrated even greater similarity to ground truth responses. In the second experiment, both SFT and DPO models generated responses more aligned with ground truth when provided with original target values rather than random ones, with the DPO model achieving superior performance. These findings suggest that models trained in single-turn settings effectively integrate target values into their responses, capturing key messages in Reddit comments and reflecting variations in target values.

In contrast, in the multi-turn setting, while the DPO model performed well in the second experiment, it did not surpass the baseline in the first experiment. This may be attributed to the increased complexity of interactions in longer threads, where it becomes challenging to identify how specific comments influence target values. For instance, even if the OP expressed positive values in their final comment, it is unclear which prior interaction contributed to this outcome. The single-turn setting simplifies these relational dynamics, making interactions more explicit. Consequently, the model trained in the single-turn setting was selected as the final reference generator.

\begin{table}[t]
\centering
\resizebox{\columnwidth}{!}{%
\begin{tabular}{@{}lcccc@{}}
\toprule
\textbf{} & \multicolumn{2}{c}{\textbf{Comparison 1}} & \multicolumn{2}{c}{\textbf{Comparison 2}} \\ 
\cmidrule(lr){2-3} \cmidrule(lr){4-5}
\textbf{} & \textbf{Llama} & \textbf{RG (SFT)} & \textbf{RG (SFT)} & \textbf{RG (DPO)} \\ \midrule
\textbf{Single-turn} & & & & \\
\hspace{2mm}Order 1 & 480 & \textbf{520} & 449 & \textbf{551} \\
\hspace{2mm}Order 2 & 368 & \textbf{632} & 495 & \textbf{505} \\ \midrule
\textbf{Multi-turn} & & & & \\
\hspace{2mm}Order 1 & \textbf{634} & 366 & \textbf{513} & 487 \\
\hspace{2mm}Order 2 & \textbf{549} & 451 & \textbf{506} & 494 \\ \bottomrule
\end{tabular}%
}
\caption{Pairwise comparison results for single-turn and multi-turn settings, evaluating the similarity of the reference generator (RG) responses to ground truth comments.}
\label{table:results-rg-exp1}
\end{table}

\begin{table}[t]
\centering
\resizebox{\columnwidth}{!}{%
\begin{tabular}{@{}lcccc@{}}
\toprule
\textbf{} & \multicolumn{2}{c}{\textbf{RG (SFT)}} & \multicolumn{2}{c}{\textbf{RG (DPO)}} \\ 
\cmidrule(lr){2-3} \cmidrule(lr){4-5}
\textbf{} & \textbf{Original} & \textbf{Random} & \textbf{Original} & \textbf{Random} \\ \midrule
\textbf{Single-turn} & & & & \\
\hspace{2mm}Order 1 & \textbf{525} & 475 & \textbf{655} & 345 \\
\hspace{2mm}Order 2 & \textbf{554} & 446 & \textbf{687} & 313 \\ \midrule
\textbf{Multi-turn} & & & & \\
\hspace{2mm}Order 1 & \textbf{584} & 416 & \textbf{751} & 249 \\
\hspace{2mm}Order 2 & 438 & \textbf{562} & \textbf{755} & 245 \\ \bottomrule
\end{tabular}%
}
\caption{Pairwise comparison results for single-turn and multi-turn settings, evaluating the performance of reference gesponse (RG) under original and random target values.}
\label{table:results-rg-exp2}
\end{table}

\subsection{Supporter Model}
\label{sec:appendix-sm}

The supporter model's training consists of two stages: SFT and DPO, with training data generated through seeker simulator simulations. The model takes the dialogue history, target values, and reference response as input, with detailed prompts provided in Table~\ref{table:prompt-supporter}. Training was performed on an NVIDIA A100-80GB GPU, with durations of approximately 20 hours for the SFT stage and 5 hours for the DPO stage.

\paragraph{SFT Stage.} During SFT, to mitigate GPT's inherent bias toward utilizing reference responses, the model generates alternative responses by reversing the decision regarding reference response usage. Specifically, the model is prompted to reverse its decision regarding the use of the reference response from the previous answer and to regenerate both Step 3 and Step 4. The overview of the SFT dataset and distribution of selected strategies are presented in Table~\ref{table:sm-sft-dataset} and Table~\ref{table:sm-sft-strategies}, respectively.

\paragraph{DPO Stage.} During DPO training, simulations are conducted to generate preference data. The supporter model generates two responses per turn: one based on its initial reference usage and another taking the opposite approach (alternative response). Each response undergoes independent simulations, and its effectiveness in reinforcing target values is quantified using a reward function (Equation~\ref{equation:dpo-reward}). The response with the higher cumulative reward is selected as the chosen response, while the other is designated as rejected.

When GPT is used as the supporter, the same prompts from the SFT stage are applied. For the supporter model (SFT), the model first generates an initial response. Subsequently, by reversing the decision on the use of the reference response from Step 3 (\textit{e.g. Yes → No}), an alternative response is generated. An overview of the DPO dataset is provided in Table~\ref{table:sm-dpo-dataset}.

\subsection{Terms and License}
We utilized Llama-3-8B-Instruct as the base model for the target value detector, reference generator, and supporter model. This model is licensed under the Llama 3 Community License Agreement. All artifacts used in this study are confirmed to be accessible for research purposes.

\begin{table}[t]
\centering
\resizebox{0.9\columnwidth}{!}{%
\begin{tabular}{@{}lcccc@{}}
\toprule
\textbf{Split} & 
\makecell{\textbf{Total}\\\textbf{Dialogues}} & 
\makecell{\textbf{Total}\\\textbf{Turns}} & 
\makecell{\textbf{Dataset}} & 
\makecell{\textbf{Dataset}\\\textit{(Filtered)}} \\ \midrule
Train & \hfill 1,796 & \hfill 16,588 & \hfill 33,176 & \hfill 33,130 \\
Dev   & \hfill 120   & \hfill 1,184  & \hfill 2,374  & \hfill 2,367  \\ \bottomrule
\end{tabular}%
}
\caption{Overview of the SFT dataset used for training the supporter model. The filtered dataset excludes instances where the generated strategy deviates from the requested strategy.}
\label{table:sm-sft-dataset}
\end{table}

\begin{table}[t]
\centering
\resizebox{\columnwidth}{!}{%
\begin{tabular}{@{}lrrrr@{}}
\toprule
\textbf{Category} & 
\makecell{\textbf{Initial}\\\textbf{Response}} & 
\makecell{\%} & 
\makecell{\textbf{Alternative}\\\textbf{Response}} & 
\makecell{\%} \\ \midrule
Question         & \hfill 393  & \hfill 2.4  & \hfill 152  & \hfill 0.9  \\
Restatement      & \hfill 1,234 & \hfill 7.4  & \hfill 723  & \hfill 4.4  \\
Reflection       & \hfill 1,251 & \hfill 7.6  & \hfill 466  & \hfill 2.8  \\
Self-disclosure  & \hfill 949  & \hfill 5.7  & \hfill \textbf{7,949} & \hfill \textbf{48.0} \\
Affirmation      & \hfill \textbf{6,577} & \hfill \textbf{39.7} & \hfill 1,014 & \hfill 6.1  \\
Suggestions      & \hfill \underline{6,159} & \hfill \underline{37.2} & \hfill \underline{5,773} & \hfill \underline{34.9} \\
Information      & \hfill 4    & \hfill 0.0  & \hfill 380  & \hfill 2.3  \\
Others           & \hfill 0    & \hfill 0.0  & \hfill 106  & \hfill 0.6  \\ \bottomrule
\end{tabular}%
}
\caption{Strategy distribution across initial and alternative responses in the supporter model's SFT training dataset.}
\label{table:sm-sft-strategies}
\end{table}

\section{Seeker Simulator}
\label{sec:ref-seeker-simulator}

\subsection{Persona Generation}
\label{sec:seeker-persona}

We develop a diverse set of seeker personas to train the supporter model, enabling it to effectively understand and address various problem scenarios. The creation of these seeker personas involves a 5 step process.

\paragraph{Step 1. Situation Generation}

\noindent We aim to create a diverse set of situations reflecting specific circumstances individuals face, each expressed in a single sentence (\textit{e.g., ``I just moved in this week, and it's so hard to make friends''}). To achieve this, we first define 6 primary problem categories and 27 subcategories based on prior research related to emotional support datasets and seeker simulator implementation \citep{liu-etal-2021-towards, zheng-etal-2024-self, lee-etal-2024-cactus, zhao-etal-2024-esc}, as detailed in Table~\ref{table:seeker-problems}.

To ensure the situations also reflect diverse human values, we integrate information about 20 distinct values. For each combination of the six problem categories and 20 values, we generate 10 to 30 unique situations using GPT-4o. This process result in a total of 2,940 unique situations. The prompts used for this process are detailed in Table~\ref{table:persona-gen-prompt1}.

\paragraph{Step 2. Evaluation on Value-Alignment}

\noindent We evaluate the alignment of the generated situations with the provided values using GPT-4o, employing a 5-point scale. Situations scoring 3 or belowa are excluded from further consideration, resulting in the retention of 2,036 situations. The evaluation prompt used for this process is detailed in Table~\ref{table:persona-gen-prompt2}.

\paragraph{Step 3. Emotion labeling}

\noindent Emotion labeling is conducted for the previously generated situations using 10 negative emotions (\textit{Frustration, Anxiety, Sadness, Fear, Guilt, Shame, Anger, Depression, Jealousy, Disgust}) identified from prior research \citep{liu-etal-2021-towards, rashkin-etal-2019-towards}. Each situation is labeled five times using GPT-4o-mini, and the final classification is determined by majority vote.

\paragraph{Step 4. Create Demographic Information}

\noindent To ensure consistency in responses generated by the seeker simulator and to enable the supporter model to interact with seekers with diverse characteristics, we generate demographic profiles including age, gender, and occupation for each simulated situation.

Our persona generation process resulted in 2,036 unique personas, each defined by problem category, situations, emotion types, and demographic information. These personas are divided into three datasets: a training set containing 1,796 personas, and development and test sets with 120 personas each. The training and development sets are used to construct SFT and DPO datasets for the supporter model through simulation, while the test set is reserved for comparative performance evaluation across models. Examples of the generated personas are provided in Table~\ref{table:persona-examples}.

\subsection{Evaluation of Seeker Simulator Performance on ESConv}
\label{sec:seeker-model}

\paragraph{Comparison Models.} Developing a supporter model capable of effectively assisting in real conversations with human seekers requires a seeker simulator that exhibits human-like behavior. To identify the most suitable model for this purpose, we conduct experiments on a range of candidates. The evaluated models are as follows:

\begin{itemize}[topsep=5pt, partopsep=0pt, itemsep=3pt, parsep=0pt]
    \item \textbf{Prompt-based models:} GPT-4o-mini, \textit{Llama-3-8B-Instruct}
    \item \textbf{Fine-tuned models:} Llama-ESConv, Llama-ExTES
    \item \textbf{Pre-existing seeker simulator:} ESC-Role \citep{zhao-etal-2024-esc}
\end{itemize}

Llama-ESConv and Llama-ExTES are fine-tuned versions of Llama-3-8B-Instruct. These models are trained on seeker turns from the ESConv dataset \citep{liu-etal-2021-towards} and the ExTES dataset \citep{zheng-etal-2024-self}, respectively.

\paragraph{Evaluation Approach.} We evaluate the models on the ESConv test dataset by providing dialogue context up to each seeker turn and generating the subsequent utterance. The evaluation compare generated responses to actual seeker utterances across four dimensions: length, content, emotions, and values.

For length, we calculate the correlation between the lengths of the generated and actual utterances. Content evaluation employs BERT-Score \footnote{\url{https://huggingface.co/sentence-transformers/all-mpnet-base-v2}} and GPT-4o-mini to assess semantic similarity between generated and reference responses. Emotional analysis uses EmoLlama-Chat-7B \citep{liu2024emollms} to determine sentiment polarity for each turn, measuring the correlation between generated and actual sentiment levels. To assess value alignment, we employ the model proposed by \citet{schroter-etal-2023-adam} to generate probability distributions across 20 values. We then calculate cosine similarity and Euclidean distance between the generated and actual distributions, reporting the mean values across all turns.

\paragraph{Results.} The experimental results are summarized in Table~\ref{table:seeker-performance}. In the ESConv test dataset, the average length of seeker utterances is 19.5, with GPT-4o-mini and Llama-ExTES exhibiting similar utterance lengths. While individual evaluation metrics show some variation, GPT-4o-mini with one-shot dialogue examples demonstrates strong overall performance. Therefore, GPT-4o-mini (one-shot) is selected as the final seeker simulator.

\subsection{Human Evaluation of Seeker Simulator}
\label{sec:seeker-human-eval}

We conducted a human evaluation to assess the naturalness of responses generated by our seeker simulator compared to real human seekers. The evaluation utilized two psychotherapy datasets: (1) ESConv \citep{liu-etal-2021-towards}, a crowdsourced emotional support dialogue dataset, and (2) AnnoMI \citep{annomi-2022, annomi-2023}, which comprises real counseling conversations from YouTube and Vimeo videos.

The evaluation sample included 200 dialogues (140 from ESConv and 60 from AnnoMI) truncated to various lengths. For each dialogue history, we compared the seeker simulator’s generated responses with the original seeker responses in a pairwise manner.

The evaluation was conducted by 16 evaluators, comprising undergraduate and graduate students from diverse academic backgrounds, including psychology, education, and computer science. To ensure an unbiased assessment, they had no prior exposure to our system. The evaluators were asked, \textit{``Which response is more natural for the seeker?''} and instructed to choose between the two responses or select ``Tie.'' While the evaluators were not psychotherapy experts, they could reliably assess response naturalness based on their own experiences with help-seeking situations, as emotional distress is a universal human experience.

The results, presented in Figure~\ref{fig:seeker-human-eval}, show that the seeker simulator's responses were rated as more natural than the original seeker responses in 66.4\% and 33.3\% of cases across the two datasets. Including ties, these percentages increased to 80.7\% and 75.0\%. Some evaluators noted that the seeker simulator effectively conveyed negative emotions and demonstrated strong situational engagement. These findings suggest that the seeker simulator achieves a level of naturalness comparable to that of real seekers.

\subsection{Prompts for Seeker Simulator}
\label{sec:seeker-prompt}

The seeker simulator generates subsequent seeker responses by integrating persona details and dialogue context. Each simulation starts with the predefined situation in the persona as the initial seeker response. A detailed prompt for the seeker simulator is presented in Table~\ref{table:seeker-prompt}.

\section{Evaluation Metrics}
\label{sec:eval-metrics}

\subsection{ES-Skills}
\label{sec:es-skills}

The definitions of the evaluation criteria for ES-Skills are as follows:

\paragraph{Emotional Support Skills}

\begin{itemize}[topsep=5pt, partopsep=0pt, itemsep=3pt, parsep=0pt]
    \item \textbf{Identification:} How effectively does the therapist explore the patient’s situation to identify underlying issues?
    \item \textbf{Comforting:} How well does the therapist demonstrate appropriate emotional responses, such as warmth, empathy, and compassion?
    \item \textbf{Suggestions:} How useful and relevant are the therapist’s suggestions for addressing the patient’s problems?
    \item \textbf{Experience:} How well does the therapist draw on their own relevant experiences to connect with the user’s situation?
    \item \textbf{Informativeness:} How specific and informative are the therapist’s responses in addressing the patient’s situation?
\end{itemize}

\paragraph{General Conversation Skills}

\begin{itemize}[topsep=5pt, partopsep=0pt, itemsep=3pt, parsep=0pt]
    \item \textbf{Consistency:} How logically structured and contextually appropriate are the therapist’s responses?
    \item \textbf{Role-adherence:} How consistently does the therapist adhere to their role, maintaining a non-contradictory and reliable approach?
    \item \textbf{Expression:} How diverse are the therapist’s conversational expressions, including the variety and creativity in language and content used?
    \item \textbf{Humanness:} How human-like and natural do the therapist’s responses sound?
\end{itemize}

\paragraph{Overall}

\begin{itemize}[topsep=5pt, partopsep=0pt, itemsep=3pt, parsep=0pt]
    \item \textbf{Overall:} How well does the therapist provide overall emotional support to the patient?
\end{itemize}

\subsection{ES-Intensity}
\label{sec:es-intensity}

This model predicts the seeker's emotional intensity after a conversation on a 5-point scale, where a lower score indicates a significant reduction in negative emotions. We applied zero-shot/few-shot prompting and fine-tuning to four different models and compared their performance using the ESConv test dataset. The final model is GPT-4o-mini (zero-shot), as it showed the highest correlation with the ground truth final emotional intensity. The results and evaluation prompts are presented in Table~\ref{tab:esc-eval-results} and Table~\ref{table:es-intensity-prompt}.

\begin{table}[!t]
    \centering
    \resizebox{\linewidth}{!}{ 
    \begin{tabular}{lccccc}
        \toprule
        \textbf{Model} & \textbf{Method} & \textbf{Acc.}↑ & \textbf{F1}↑ & \textbf{MSE}↓ & \textbf{Corr.}↑ \\ 
        \midrule
        Baseline & - & 0.435 & 0.264 & \textbf{0.768} & - \\ 
        \midrule
        GPT-4o & Zero-shot & 0.358 & 0.352 & 1.182 & 0.303 \\ 
         & Few-shot & 0.415 & 0.416 & 1.057 & 0.312 \\ 
        \midrule
        GPT-4o-mini & Zero-shot & \textbf{0.466} & \textbf{0.432} & 0.875 & \textbf{0.345} \\ 
         & Few-shot & 0.415 & 0.410 & 0.966 & 0.327 \\ 
        \midrule
        Llama3-8B & Zero-shot & 0.426 & 0.318 & 0.869 & 0.130 \\ 
         & Fine-tuned & 0.409 & 0.395 & 0.892 & 0.330 \\ 
        \midrule
        EmoLlama-7B & Zero-shot & 0.384 & 0.289 & 0.972 & 0.084 \\ 
         & Fine-tuned & 0.407 & 0.373 & 0.977 & 0.185 \\ 
        \bottomrule
    \end{tabular}}
    \caption{Evaluation results of different models on final emotional intensity prediction tasks. The metrics are accuracy, weighted F1-score, mean squared error, and Spearman's correlation coefficient. The baseline model predicts all final emotional intensities as 2.}
    \label{tab:esc-eval-results}
\end{table}

\subsection{ES-Value}
\label{sec:es-value}

To evaluate the effectiveness of value reinforcement, it is essential to consider two perspectives: the seeker’s and the supporter’s. These viewpoints provide a comprehensive understanding of how effectively positive values are identified, discussed, and integrated into the seeker’s mindset during emotional support conversations. The definitions for each perspective are as follows:

\begin{itemize}
    \item \textbf{Seeker's perspective:} How strongly were positive human values explored and reinforced in the patient through the conversation?
    \item \textbf{Supporter's perspective:} How effectively did the therapist help the patient in exploring and reinforcing positive human values?
\end{itemize}

ES-Value is assessed through pairwise comparisons between a reference model and multiple baseline models. Dialogues from the reference model are paired with corresponding dialogues from baseline models, ensuring the seeker personas are identical. Each pair is evaluated 10 times, and the reference model's win ratio is normalized to a score ranging from 0 to 1. This evaluation utilize GPT-4o-mini as the assessment model (see Table~\ref{table:es-value-prompt} for prompt details).

\section{Descriptions of Emotional Support Methods Selected as Baselines}
\label{sec:es-method-driven}

We select baselines from methods that have contributed to the development of supporter models, ensuring diverse characteristics. The explanations for each method are as follows:

\begin{itemize}
    \item \textbf{Ask-an-Expert} \citep{zhang-etal-2023-ask}: This approach involves consulting an expert-role LLM at every turn of the conversation to obtain advice on the seeker’s emotional status, its cause, and potential solutions, which are then leveraged to generate the supporter’s response.
    \item \textbf{ESCoT} \citep{zhang-etal-2024-escot}: This method follows a chain-of-thought process, considering the seeker's emotional state, emotion stimulus (the specific trigger of the emotion), the seeker’s personal interpretation of the stimulus, and the reasoning behind the selected support strategy before generating a supporter response.
    \item \textbf{PPDPP} \citep{deng2024ppdpp}: This approach involves training a policy planner to select the optimal support strategy through two steps: (1) fine-tuning on the ESConv dataset and (2) simulating and evaluating diverse conversations using three LLMs (supporter LLM, seeker LLM, and reward LLM), followed by reinforcement learning based on the rewards.
\end{itemize}

\section{Training Details for Emotion-Reinforced Models}
\label{sec:emo-reinforced}

This study investigates whether reinforcing values, rather than positive emotions, leads to more effective emotional support. To test this hypothesis, we adapt our methods by modifying the learning objective to prioritize promoting positive emotions in seekers. This approach requires two key components: a reference generator and a supporter model, both optimized for emotional reinforcement. Unlike the value-based method, this approach does not require a target value detector. The following subsections outline the training procedures for the reference generator and the supporter model.

\subsection{Reference Generator}
The reference generator is trained on supporter response from Reddit that successfully elicited positive emotional responses from OPs. This training approach ensures that the generated responses effectively foster positive emotions. Given the dialogue history (\(o_1, c_1, o_2, c_2, ..., o_t\)), the model generates a supporter response \((c_t)\) as follows:
\begin{equation}
c_t = \text{LLM}_\text{RG}(o_1, c_1, o_2, c_2, ..., o_t)    
\end{equation}

Unlike our model, which incorporates both dialogue history and target values, this generator relies solely on dialogue history as input. Therefore, it employs only the SFT stage. Although the training data and prompts are consistent with those used for our model, all value-related information has been excluded from the reference generator's training process.

\subsection{Supporter Model}

The supporter model for emotion reinforcement processes two inputs: the dialogue history and a reference response generated by the reference generator. At each turn, the model performs reasoning across four key aspects: (1) identifying the seeker’s issues and current emotional state, (2) analyzing the content of the reference response, (3) deciding whether to integrate the reference response, and (4) selecting an optimal emotional support strategy to generate the subsequent response. These reasoning aspects are identical to those used in our model.

The training process for the supporter model involves both SFT and DPO using data generated through simulations with a seeker simulator based on GPT-4o-mini.

\paragraph{SFT Stage.} Similar to our approach, a dual-generation method is employed: GPT produces two responses per turn---one with references and one without---ensuring balanced training data within identical contexts. The simulation-generated data is then used to fine-tune Llama-3-8B-Instruct, with dataset sizes detailed in Table~\ref{table:emo-sm-dataset-size}.

\begin{table}[t]
\centering
\resizebox{0.65\columnwidth}{!}{%
\begin{tabular}{@{}llrr@{}}
\toprule
\textbf{Stage} & \textbf{Supporter} & \textbf{Train} & \textbf{Dev} \\ \midrule
SFT & GPT-4o-mini & 24,580 & 1,656 \\
DPO & GPT-4o-mini & 2,610 & 552 \\
\bottomrule
\end{tabular}%
}
\caption{Dataset sizes for training the supporter model for positive emotion reinforcement generated through simulation. The `Supporter' column refers to the supporter model used in the simulation.}
\label{table:emo-sm-dataset-size}
\end{table}

\paragraph{DPO Stage.} This stage optimizes the supporter model to generate responses that more effectively promote positive emotions. The process uses simulated dialogues between the supporter model and a seeker simulator to generate training data. For each turn, the supporter model produces two responses: one following its initial reference usage and another taking the opposite approach. Both responses undergo simulation to evaluate their emotional impact, with the more effective response marked as preferred. The cumulative reward for a supporter's response at turn \(t\) (\(u_t^\text{sup}\)) is calculated as:
\begin{equation}
R(u_t^\text{sup}) = \sum_{k=1}^{h} \gamma^{k-1} S_{t+k}(u_t^\text{sup})    
\end{equation}

where \(S_t(u_t^\text{sup})\) represents the emotion score at turn \(t\) calculated by GPT-4o-mini, \(h\) is the look-ahead horizon (the number of future steps considered), and \( \gamma \) is a discount factor balancing immediate and future rewards. Response pairs are included in the DPO dataset when their reward difference exceeds the threshold \(T_\text{diff}\). 

Emotion scores are calculated using prompts inspired by \citet{deng2024ppdpp}. For each turn, GPT evaluates the seeker's emotional state as ``feels worse'', ``feels the same'', ``feels better'', or ``the issue has been solved''. These responses are then mapped to scores of -1.0, -0.5, 0.5, and 1.0, respectively. Ten responses are collected for each turn, and the average score is used as the final emotion score. The prompts used for this process are provided in Table~\ref{table:emo_score-prompt}.

These simulations use GPT-4o-mini, and dataset sizes are summarized in Table~\ref{table:emo-sm-dataset-size}.

\section{Evaluating the Success of Target Values Reinforcement}
\label{sec:value-reinforcement}

We engaged each model in conversations with a seeker simulator and analyzed the frequency of target values appearing in the seeker's subsequent responses. The success rate was assessed based on valid turns---the number of future turns within which the target values remained relevant. For example, if the valid turn threshold was three, reinforcement was considered successful if the seeker's response included the target values within the next three turns.
To control for variations in dialogue length, we considered only cases where at least one positive seeker response occurred within the valid turns. This approach mitigated the influence of longer conversations, where negative responses might become more frequent and lower the success rate.

\section{Training Details for Generalization Capability Evaluation}
\label{sec:generalization}

To evaluate the generalization capability of our method, we conducted experiments using the Cactus dataset \citep{lee-etal-2024-cactus} instead of Reddit. Cactus is a counseling dataset based on \textit{Cognitive Behavioral Therapy (CBT)}, a therapeutic approach that helps individuals identify and modify negative thought patterns and behaviors. The conversations were generated using GPT-4o and validated through human evaluation to ensure their suitability as psychological counseling dialogues. 

We utilized Cactus to train both the target value detector and the reference generator. Subsequently, we employed these models in a simulation to create SFT and DPO datasets, which were then used to train the supporter model. The details for each stage are outlined below.

\subsection{Target Value Detector \& Reference Generator}

The Cactus dataset contains a total of 31,577 counseling dialogues. To consider resource efficiency, we selected 4,057 dialogues by excluding cases where the seeker's intake form---including personal information, issues, history, and other details---was duplicated across conversations. As described in Section~\ref{sec:reddit-data}, we then used models developed by \citet{liu2024emollms} and \citet{schroter-etal-2023-adam} to label each seeker’s turn with sentiment strength and expressed values. These results were subsequently used to prepare the dataset for training the target value detector and reference generator. The final dataset consists of 11,000 instances for training and 1,631 instances each for validation and testing.

Following the approach outlined in Section~\ref{sec:method}, we trained the target value detector and reference generator separately. Unlike Reddit, where a single comment can receive multiple replies---leading to a branching conversation structure---the Cactus dataset follows a unidirectional conversational flow. Due to this structure, the DPO process was omitted during reference generator training.

\begin{table}[t]
\centering
\resizebox{0.65\columnwidth}{!}{%
\begin{tabular}{@{}llrr@{}}
\toprule
\textbf{Stage} & \textbf{Supporter} & \textbf{Train} & \textbf{Dev} \\ \midrule
SFT & GPT-4o-mini & 38,543 & 2,640 \\
DPO & GPT-4o-mini & 1,270 & 256 \\
\bottomrule
\end{tabular}%
}
\caption{Dataset sizes for supporter model training, generated through simulation using Cactus-based target value detector and reference generator. The `Supporter' column refers to the supporter model used in the simulation.}
\label{table:generalization-dataset-size}
\end{table}

\subsection{Supporter Model}
We trained the supporter model in two stages, SFT and DPO, following the approach introduced in Section~\ref{sec:supporter}. 
The training dataset was generated through interactions between GPT-4o-mini-based supporter and seeker simulators. 
The key difference from Section~\ref{sec:supporter} is that the target value detector and reference generator used for simulation were trained on the Cactus dataset instead of Reddit. 
Additionally, when constructing the dataset for DPO, we adopted the hyperparameter settings (\(h=3\), \( \gamma=1 \),  \(T_\text{diff}=2\)) that achieved the best performance on the Reddit dataset. 
Detailed information on dataset size is provided in Table~\ref{table:generalization-dataset-size}.

\begin{table}[t]
\centering
\resizebox{0.7\columnwidth}{!}{%
\begin{tabular}{@{}llp{1.2cm}@{}}
\toprule
\hspace{2mm}\textbf{Category} & \textbf{Metric} & \textbf{Corr.} \\ 
\midrule
\hspace{2mm}ES-Skills & Identification & 0.422$^{*}$ \\
& Comforting & 0.322$^{*}$ \\
& Suggestions & 0.421$^{*}$ \\
& Experience & 0.778$^{*}$ \\
& Informativeness & 0.282$^{*}$ \\
& Consistency & 0.351$^{*}$ \\
& Role-Adherence & 0.235$^{\dagger}$ \\
& Expression & 0.198 \\
& Humanness & 0.202 \\
& Overall & 0.413$^{*}$ \\
\midrule
\hspace{2mm}ES-Value & Seeker & 0.332$^{*}$ \\
& Supporter & 0.413$^{*}$ \\
\bottomrule
\end{tabular}%
}
\caption{Spearman's rank correlation between expert and GPT-generated scores. Significant correlations are marked with * ($p$-value $< 0.05$) and $\dagger$ ($p$-value $< 0.1$).}
\label{table:expert-corr}
\end{table}

\section{Validation of Automated Evaluation Models}
\label{sec:appendix-expert-corr}
To evaluate the performance of GPT-4o-mini in assessing ES-Skills and ES-Value, we analyzed the correlation between expert evaluation scores and GPT-generated scores. For this purpose, we randomly select and evaluate 60 dialogues generated by our models and baselines. For ES-Value, we compared individual dialogue scores generated by GPT and expert evaluations, rather than using a pairwise scoring approach, to enable a more straightforward comparison. As shown in Table~\ref{table:expert-corr}, significant correlations were observed across most metrics, except for \textit{Expression} and \textit{Humanness}. These findings suggest that automated evaluation models can reliably approximate human assessments of emotional support, conversational quality, and value reinforcement, supporting the validity of our experimental results.

\section{Details of Experiments on Self-Disclosure}
\label{sec:appendix-self-disclosure}

Self-disclosure---sharing experiences related to those of the seeker---is a key strategy in emotional support for fostering intimacy and reducing stress \citep{10.1093/scan/nsae003, 10.1093/jcmc/zmab005}.
The ability to share such experiences has been used as an evaluation criterion for emotional support systems \citep{zhang-etal-2023-ask}.
However, some users might feel uncomfortable when dialogue systems present these experiences as personal.

To better understand the impact of self-disclosure on emotional support systems, we investigated two alternative approaches: (1) removing it entirely and substituting the next most probable strategy in the supporter's reasoning process (Section~\ref{sec:supporter}); and (2) rephrasing self-disclosure responses to frame them as experiences of others, using GPT-4o-mini.

As shown in Table~\ref{table:exp3-no-sd1} and Table~\ref{table:exp3-no-sd2}, the models consistently exhibited declines in overall ES-Skills and ES-Value when self-disclosure was modified or removed. The metric most affected was \textit{Experience}, with related metrics such as \textit{Suggestions} and \textit{Informativeness} also showing performance drops.

The results reinforces the importance of self-disclosure in emotional support but, at the same time, highlight the need for research on more sophisticated methods for experience sharing.
For example, analyzing different types of self-disclosure and developing alternative strategies based on seeker perceptions could offer meaningful improvements. This detailed investigation will be left for future work.

\begin{table}[t]
\centering
\resizebox{\columnwidth}{!}{%
\begin{tabular}{@{}cc@{}}
\toprule
\hspace{2mm}\textbf{Category} & \textbf{Freq.} \\ 
\midrule
\hspace{2mm}Strengths and Achievements Acknowledgment & 30 \\
\hspace{2mm}Exploration of Issues and Challenges & 23 \\
\hspace{2mm}Self-Compassion and Acceptance & 19 \\
\hspace{2mm}Exploration of Personal Interests & 16 \\
\hspace{2mm}Emotional Resilience and Coping Strategies & 14 \\
\hspace{2mm}Exploration of Goals and Motivations & 14 \\
\hspace{2mm}Motivation and Alignment with Goals & 1 \\
\bottomrule
\end{tabular}%
}
\caption{Frequency of areas for improvement in value reinforcement.}
\label{table:value-issue-freq}
\end{table}

\section{Detailed Analysis of Value Reinforcement Performance}
\label{sec:appendix-indepth-value}
To identify areas for improvement in our DPO model's value reinforcement, we conducted a detailed analysis. First, we evaluated value reinforcement scores from both seeker and supporter perspectives using GPT-4o-mini on a 5-point scale. Next, we analyzed 40 dialogues that received the lowest score of 4 to identify potential improvement areas. This analysis involved reasoning with GPT-4o-mini and categorizing the areas, as summarized in Table~\ref{table:value-issue-category}. Subsequently, GPT-4o-mini was used again to assign up to three relevant categories to each of the 40 dialogues. The frequency of issues for each category is detailed in Table~\ref{table:value-issue-freq}.

\begin{table*}[t]
\centering
\resizebox{\textwidth}{!}{%
\begin{tabular}{@{}p{2cm}p{16cm}@{}}
\toprule
\textbf{\hspace{2mm}System} & Select and return up to 3 values to reinforce in the patient for effective emotional support. \\ \midrule
\textbf{\hspace{2mm}User} & Human values: \textit{\{List of 20 human values\}}
\newline\newline The dialogue history below is a conversation between a patient experiencing emotional difficulties and a therapist providing support. For effective emotional support, which values should be reinforced in the patient so that they are expressed more frequently in the future? Select up to 3 values from the list provided above. Answer in the format `value1, value2, value3' separated by commas without any additional explanation.
\newline\newline Dialogue history: \textit{\{Dialogue history\}} \\ \bottomrule
\end{tabular}%
}
\caption{Training prompts for the target value detector.}
\label{table:prompt-tvd}
\end{table*}

\begin{table*}[t]
\centering
\resizebox{\textwidth}{!}{%
\begin{tabular}{@{}p{2cm}p{16cm}@{}}
\toprule
\textbf{\hspace{2mm}System} & You will take on the role of a therapist to help a patient with emotional difficulties, aiming to reduce their distress and support them in overcoming their challenges. \\ \midrule
\textbf{\hspace{2mm}User} & 1. Dialogue history: \textit{\{Thread history\}}\newline\newline 
2. Target values:\newline
\textit{\{Information on the target values\}}\newline\newline 
As a therapist supporting a patient with emotional difficulties, your goal is to reduce their distress and guide them through challenges. The target values are those that are expected to be more frequently expressed by the patient. Generate the next turn of the utterance based on the dialogue history, aiming to reinforce these target values in the patient. \\ \bottomrule
\end{tabular}%
}
\caption{Training prompts for the reference generator. The target values information includes the definition of each value along with the set of contained values.}
\label{table:prompt-rg}
\end{table*}

\begin{table*}[t]
\centering
\resizebox{\textwidth}{!}{%
\begin{tabular}{@{}p{2cm}p{16cm}@{}}
\toprule
\textbf{\hspace{2mm}System} & Determine which of the two comments generated by each model is more similar to the ground truth comment. \\ \midrule
\textbf{\hspace{2mm}User} & Thread: \textit{\{Thread history\}} \newline\newline
Ground truth comment: \textit{\{GT comment\}} \newline\newline
The above includes the thread history and the corresponding ground truth comment, which continues the thread. Below are two comments generated by different models. First, provide reasoning for your evaluation, and then select the comment that is more similar in content to the ground truth comment. \newline\newline
Comment A: \textit{\{Response from model A\}} \newline
Comment B: \textit{\{Response from model B\}} \newline\newline
[Template] \newline
Reasoning: \newline
Similar comment: Answer with either 'Comment A' or 'Comment B' only \\ \bottomrule
\end{tabular}%
}
\caption{Prompts for evaluating the performance of the reference response.}
\label{table:prompt-rg-evaluation}
\end{table*}

\begin{table*}[t]
\centering
\resizebox{\textwidth}{!}{%
\begin{tabular}{@{}p{2cm}p{16cm}@{}}
\toprule
\textbf{\hspace{2mm}System} & You will take on the role of a therapist to help a patient with emotional difficulties, aiming to reduce their distress and support them in overcoming their challenges. \\ \midrule
\textbf{\hspace{2mm}User} & 1. Strategies for emotional support: \textit{\{Definition of 8 emotional support strategies\}} \newline
2. Dialogue history: \textit{\{Dialogue history\}} \newline
3. Target values: \textit{\{Information on the target values\}} \newline
4. Reference response: \textit{\{Reference response\}} \newline

As a therapist supporting a patient with emotional difficulties, your goal is to reduce their distress and guide them through challenges. The target values are those that are expected to be more frequently expressed by the patient. You need to generate the therapist's next utterance based on the dialogue history, aiming to reinforce these target values in the patient. \newline

The therapist’s next utterance should follow these guidelines: \newline
- Use only one sentence without any extra explanation, framing, introductory phrases, or meta-commentary \newline
- Avoid directly mentioning the target values, but focus on reinforcing them through your guidance. \newline
- If the patient shows signs of improvement in the dialogue history, acknowledge their progress and guide the conversation to an efficient close. \newline
- Do not repeat similar messages from previous therapist utterances in the dialogue history. \newline

The reference response is a therapist's reply given to another patient in a similar situation, which you can use as a reference for generating your next response. Before generating the therapist’s response to satisfy the above conditions, thoroughly analyze the following:\newline
Step 1. Understanding the patient's issues and current state \newline
- What is the patient's issue? \newline
- Have their situation and the causes of their emotions been sufficiently explored? If not, what additional information should be obtained to deeply understand them? \newline
- What is the patient's current emotional state? How have the patient's emotions or thoughts changed through the conversation? \newline

Step 2. Identifying the key points of the reference response \newline
- What is the main message in the referenced response (item 4)? \newline

Step 3. Determination of reference response usage \newline
- Would using a reference response be helpful for generating the next therapist utterance? Why or why not? \newline
- If a reference response is used, how would it be applied, and if it is not used, what alternative message would be provided? \newline

Step 4. Therapist's next strategy and response \newline
- Based on the above (Step 1-Step3), what emotional support strategy should be used, and what message should you convey to the patient in the next response? \newline

You should respond in the following template format: \newline
Step 1. Understanding the patient's issues and current state \newline
-Reasoning: (the result of your analysis) \newline
Step 2. Identifying the key points of the reference response \newline
-Reasoning: (the result of your analysis) \newline
Step 3. Determination of reference response usage \newline
-Reasoning: (The result of your analysis, starting with whether to use the reference response --- 'Yes' or 'No') \newline
Step 4. Therapist's next strategy and response \newline
-Strategy: (choose one emotional support strategy for the next turn based on the reasoning) \newline
-Response: (only the therapist’s next utterance without any explanation) \\ \bottomrule
\end{tabular}%
}
\caption{Training prompts for the supporter model.}
\label{table:prompt-supporter}
\end{table*}

\begin{table*}[ht]
\centering
\resizebox{0.7\textwidth}{!}{%
\begin{tabular}{@{}ccccrrrrrrrrr@{}}
\toprule
\multirow{3.5}{*}{\makecell[c]{\textbf{Supporter}\\\textbf{Model}}} & 
\multirow{3.5}{*}{\makecell[c]{\textbf{\(h\)}}} & 
\multirow{3.5}{*}{\makecell[c]{\textbf{\( \gamma \)}}} & 
\multirow{3.5}{*}{\makecell[c]{\textbf{\( T_\text{diff} \)}}} & 
\multicolumn{3}{c}{\textbf{Train}} & 
\multicolumn{3}{c}{\textbf{Dev}} \\ 
\cmidrule(lr){5-7} \cmidrule(lr){8-10}
& & & & 
\multirow{2.5}{*}{\makecell[c]{\textbf{\# of}\\\textbf{Data}}} & 
\multicolumn{2}{c}{\textbf{Chosen}} & 
\multirow{2.5}{*}{\makecell[c]{\textbf{\# of}\\\textbf{Data}}} & 
\multicolumn{2}{c}{\textbf{Chosen}} \\ 
\cmidrule(lr){6-7} \cmidrule(lr){9-10}
& & & & & 
\makecell[c]{\textbf{Initial}} & 
\makecell[c]{\textbf{Alternative}} & & 
\makecell[c]{\textbf{Initial}} & 
\makecell[c]{\textbf{Alternative}} \\ 
\midrule
\textbf{GPT} & All & 1 & 1 & 2,561 & 1,168 & 855 & 458 & 208 & 149 \\
& & & 2 & 2,023 & 881 & 623 & 357 & 268 & 149 \\
\cmidrule(lr){2-10}
& All & 0.9 & 1 & 1,712 & 947 & 765 & 298 & 164 & 134 \\
& & & 1.5 & 1,345 & 735 & 610 & 229 & 117 & 112 \\
\cmidrule(lr){2-10}
& 3 & 1 & 1 & 1,796 & 1,127 & 669 & 319 & 201 & 118 \\
& & & 2 & 1,144 & 724 & 420 & 210 & 132 & 78 \\
\cmidrule(lr){2-10}
& 5 & 1 & 1 & 2,015 & 1,301 & 714 & 360 & 237 & 123 \\
& & & 2 & 1,438 & 955 & 483 & 247 & 165 & 82 \\ 
\midrule
\textbf{SFT} & All & 1 & 1 & 3,301 & 1,255 & 1,106 & 628 & 263 & 206 \\
& & & 2 & 2,361 & 920 & 800 & 469 & 182 & 149 \\
\cmidrule(lr){2-10}
& All & 0.9 & 1 & 1,975 & 979 & 996 & 407 & 209 & 198 \\
& & & 1.5 & 1,556 & 777 & 779 & 318 & 153 & 165 \\
\cmidrule(lr){2-10}
& 3 & 1 & 1 & 1,825 & 1,122 & 703 & 375 & 226 & 149 \\
& & & 2 & 1,117 & 663 & 454 & 239 & 153 & 116 \\
\cmidrule(lr){2-10}
& 5 & 1 & 1 & 2,186 & 1,360 & 826 & 456 & 281 & 175 \\
& & & 2 & 1,489 & 919 & 570 & 172 & 116 & 56 \\ 
\bottomrule
\end{tabular}%
}
\caption{Overview of the DPO dataset categorized by the supporter model used for simulations and variations in reward calculation hyperparameters (\(h\): look-ahead horizon, \( \gamma \): discount factor, \(T_\text{diff}\): difference threshold). The ``Chosen'' column represents the distribution of chosen responses selected between the model's initial and alternative outputs.}
\label{table:sm-dpo-dataset}
\end{table*}

\begin{table*}[t]
\centering
\resizebox{\textwidth}{!}{%
\begin{tabular}{@{}llccc*{10}{l}l@{}}
\toprule
\multirow{2.5}{*}{\makecell[c]{\textbf{Categories}}} & \multirow{2.5}{*}{\makecell[c]{\textbf{Models}}} & \multirow{2.5}{*}{\makecell[c]{\textbf{\(h\)}}} & \multirow{2.5}{*}{\makecell[c]{\textbf{\( \gamma \)}}} & \multirow{2.5}{*}{\makecell[c]{\textbf{\(T_\text{diff}\)}}} & 
\multicolumn{10}{c}{\textbf{ES-Skills}↑} & \multirow{2.5}{*}{\makecell[c]{\textbf{ES-}\\\textbf{Intensity}↓}} \\ \cmidrule(lr){6-15}
& & & & & \textbf{Iden.} & \textbf{Comf.} & \textbf{Sugg.} & \textbf{Expe.} & \textbf{Info.} & \textbf{Cons.} & \textbf{Role.} & \textbf{Expr.} & \textbf{Huma.} & \textbf{Over.} & \\ \midrule

\multirow{1}{*}{Prompt-Based} &
GPT & - & - & - & 4.83 & 4.92 & 4.57$^{*}$ & 3.11$^{*}$ & 4.42$^{*}$ & \textbf{5.00} & \textbf{5.00} & 4.10$^{*}$ & 4.70 & 4.72$^{*}$ & \hspace{4mm}1.89 \\
& Llama & - & - & - & 4.87 & 4.91 & 4.43$^{*}$ & 2.91$^{*}$ & 4.47 & 4.99 & \textbf{5.00} & 4.03$^{*}$ & 4.63$^{*}$ & 4.68$^{\dagger}$ & \hspace{4mm}1.99$^{\dagger}$ \\ \midrule

\multirow{1}{*}{\makecell[l]{{ES Datasets}}} &
Llama-Reddit & - & - & - & 3.38$^{*}$ & 3.74$^{*}$ & 3.21$^{*}$ & 2.59$^{*}$ & 2.99$^{*}$ & 3.94$^{*}$ & 4.35$^{*}$ & 3.37$^{*}$ & 3.81$^{*}$ & 3.40$^{*}$ & \hspace{4mm}1.97 \\
& Llama-ESConv & - & - & - & 4.35$^{*}$ & 4.43$^{*}$ & 4.06$^{*}$ & 2.65$^{*}$ & 3.88$^{*}$ & 4.82$^{*}$ & 4.97$^{*}$ & 3.79$^{*}$ & 4.25$^{*}$ & 4.22$^{*}$ & \hspace{4mm}1.87 \\
& Llama-ExTES & - & - & - & 4.83 & 4.90 & 4.53$^{*}$ & 2.71$^{*}$ & 4.44$^{*}$ & 4.99 & \textbf{5.00} & 4.02$^{*}$ & 4.59$^{*}$ & 4.66$^{*}$ & \hspace{4mm}\underline{1.67}$^{*}$ \\
& Llama-Psych8k & - & - & - & 4.84 & 4.85$^{*}$ & 4.75 & 2.89$^{*}$ & 4.63 & 4.99 & \textbf{5.00} & 4.05$^{*}$ & 4.57$^{*}$ & 4.75 & \hspace{4mm}\textbf{1.53}$^{*}$ \\ \midrule

\multirow{1}{*}{\makecell[l]{{ES Methods}}} 
& Ask-an-Expert & - & - & - & 4.13$^{*}$ & 4.30$^{*}$ & 3.93$^{*}$ & 3.12$^{*}$ & 3.70$^{*}$ & 4.61$^{*}$ & 4.91$^{*}$ & 3.74$^{*}$ & 4.21$^{*}$ & 4.08$^{*}$ & 
\hspace{4mm}1.86 
\\
& ESCoT & - & - & - & 3.69$^{*}$ & 3.91$^{*}$ & 3.16$^{*}$ & 1.81$^{*}$ & 3.07$^{*}$ & 4.16$^{*}$ & 4.81$^{*}$ & 2.95$^{*}$ & 3.64$^{*}$ & 3.51$^{*}$ & 
\hspace{4mm}2.25$^{*}$ 
\\
& PPDPP & - & - & - & 4.64$^{*}$ & 4.88 & 4.45$^{*}$ & 2.49$^{*}$ & 4.26$^{*}$ & 4.99 & \textbf{5.00} & 3.99$^{*}$ & 4.54$^{*}$ & 4.54$^{*}$ & 
\hspace{4mm}1.83 
\\
\midrule

\multirow{1}{*}{\makecell[l]{{Emotion}\\{Reinforced}}} &
SFT & - & - & - & 4.83 & 4.91 & 4.51 & 3.64 & 4.43 & 4.97 & 4.99 & 4.16 & 4.67 & 4.73 & \hspace{4mm}1.97 \\
\cmidrule(lr){2-16}
& DPO (GPT) & 3 & 1 & 0.5 & 4.85 & 4.92 & 4.74 & \textbf{4.05} & 4.61 & 4.99 & \textbf{5.00} & \textbf{4.33} & \textbf{4.78} & 4.82 & \hspace{4mm}1.86 \\
&  & 5 & 1 & 0.5 & 4.85 & \textbf{4.95} & 4.68 & 3.80 & 4.58 & 4.99 & \textbf{5.00} & 4.28 & 4.77 & 4.81 & \hspace{4mm}1.82 \\ \midrule

\multirow{1}{*}{\makecell[l]{{ES-VR}\\{(Ours)}}} &
SFT & - & - & - & 4.85 & 4.90 & 4.72 & 3.76 & 4.56 & 4.99 & \textbf{5.00} & 4.25 & 4.73 & 4.80 & \hspace{4mm}1.86 \\
\cmidrule(lr){2-16}

& DPO (GPT) & All & 1 & 1 & 4.89 & 4.92 & 4.75 & 3.71 & 4.63 & 4.99 & \textbf{5.00} & 4.24 & 4.72 & 4.80 & \hspace{4mm}1.90 \\
& & All	& 1	& 2	& 4.85 & 4.91 & 4.73 & \underline{3.89}	& 4.63$^{\dagger}$ & \textbf{5.00} &	\textbf{5.00} &	4.27 & 4.76 & 4.80 & \hspace{4mm}1.90 \\
& & All & 0.9 & 1 & 4.84 & 4.87 & 4.72 & 3.72& 4.61 & 4.98 & \textbf{5.00} & 4.23 & 4.72 & 4.78 & \hspace{4mm}1.85 \\
& & All & 0.9 & 1.5 & 4.89 & 4.93 & 4.71 & 3.79 & 4.59 & 4.99 & \textbf{5.00} & 4.28 & 4.76 & 4.83$^{*}$ & \hspace{4mm}1.88 \\
& & 3 & 1 & 1 & 4.87 & 4.90 & 4.77 & 3.58 & 4.63 & 4.99 & \textbf{5.00} & 4.26 & 4.74 & 4.79 & \hspace{4mm}1.87 \\
& & 3 & 1 & 2 & \textbf{4.91}$^{\dagger}$ & 4.93 & \underline{4.78}$^{*}$ & 3.61 & 4.65$^{\dagger}$ & 4.99 & 5.00 & 4.23 & 4.72 & \underline{4.85}$^{*}$ & \hspace{4mm}1.94 \\
& & 5 & 1 & 1 & 4.88$^{\dagger}$ & 4.93 & 4.77$^{*}$ & 3.62 & 4.64$^{\dagger}$ & \textbf{5.00} & \textbf{5.00} & 4.25 & 4.73 & 4.83 & \hspace{4mm}1.84 \\
& & 5 & 1 & 2 & 4.88 & 4.94 & 4.73 & 3.64 & 4.66$^{*}$ & \textbf{5.00} & \textbf{5.00} & 4.24 & 4.75 & 4.84$^{*}$ & \hspace{4mm}1.83 \\ 
\cmidrule(lr){2-16}

& DPO (SFT) & All & 1 & 1 & 4.83 & 4.89 & 4.71& 3.75 & 4.60 & 4.98 & 4.99$^{\dagger}$ & 4.21 & \underline{4.77} & 4.78 & \hspace{4mm}1.93 \\
& & All & 1 & 2 & 4.89 & \textbf{4.95} & 4.77 & 3.76 & \underline{4.66}$^{*}$ & 4.99 & \textbf{5.00} & 4.26 & 4.75 & 4.83 & \hspace{4mm}1.87 \\
& & All & 0.9 & 1 & 4.86 & 4.93$^{\dagger}$ & 4.72 & 3.72 & 4.64$^{\dagger}$ & 4.99 & \textbf{5.00} & 4.26 & 4.73 & 4.82$^{\dagger}$ & \hspace{4mm}1.87 \\
& & All & 0.9 & 1.5 & 4.85 & 4.92 & 4.73 & 3.79 & 4.59 & 4.99 & \textbf{5.00} & 4.28 & 4.74 & 4.81 & \hspace{4mm}1.86 \\
& & 3 & 1 & 1 & 4.86 & 4.91 & 4.71 & 3.58 & 4.57 & 4.99 & \textbf{5.00} & 4.24 & 4.71 & 4.78 & \hspace{4mm}1.89 \\
& & 3 & 1 & 2 & \underline{4.90}$^{\dagger}$ & \textbf{4.95} & \textbf{4.80}$^{*}$ & 3.85 & \textbf{4.69}$^{*}$ & \textbf{5.00} & \textbf{5.00} & \underline{4.30} & 4.77 & \textbf{4.87}$^{*}$ & \hspace{4mm}1.75 \\
& & 5 & 1 & 1 & 4.85 & 4.90 & 4.69 & 3.76 & 4.56 & 4.99 & \textbf{5.00} & 4.23 & 4.74 & 4.78 & \hspace{4mm}1.86 \\
& & 5 & 1 & 2 & 4.83 & 4.91 & 4.70 & 3.66 & 4.61 & 4.98 & \textbf{5.00} & 4.23 & 4.74 & 4.79 & \hspace{4mm}1.91 \\ \bottomrule

\end{tabular}%
}
\caption{Comparison of models based on ES-Skills and ES-Intensity. Statistically significant differences compared to our SFT model are marked with * ($p$-value $< 0.05$), and differences with $p$-value $< 0.1$ are marked with $\dagger$, as determined by the Mann-Whitney U test.}
\label{table:exp2-baseline-comparison-all}
\end{table*}

\begin{table*}[ht]
\centering
\resizebox{0.9\textwidth}{!}{%
\begin{tabular}{@{}lllll*{6}{l}@{}}
\toprule
\multirow{2.5}{*}{\makecell[c]{\hspace{2mm}\textbf{Categories}}} & \multirow{2.5}{*}{\makecell[c]{\textbf{Models}}} & \multirow{2.5}{*}{\makecell[c]{\textbf{\(h\)}}} & \multirow{2.5}{*}{\makecell[c]{\textbf{\( \gamma \)}}} & \multirow{2.5}{*}{\makecell[c]{\textbf{\(T_\text{diff}\)}}} & \multicolumn{2}{c}{\textbf{SFT}} & \multicolumn{2}{c}{\textbf{DPO (GPT)}} & \multicolumn{2}{c}{\textbf{DPO (SFT)}} \\ 
\cmidrule(lr){6-7} \cmidrule(lr){8-9} \cmidrule(lr){10-11}
& & & & & \textbf{Seeker} & \textbf{Supporter} & \textbf{Seeker} & \textbf{Supporter} & \textbf{Seeker} & \textbf{Supporter} \\ 
\midrule

\multirow{1}{*}{\hspace{2mm}Prompt-Based} & GPT & - & - & \hspace{2mm}- & \hspace{2mm}0.51 & \hspace{4mm}0.48 & \hspace{2mm}0.49$^{\dagger}$ & \hspace{4mm}0.42$^{*}$ & \hspace{2mm}0.49$^{*}$ & \hspace{4mm}0.42$^{*}$ \\ 
& Llama & - & - & \hspace{2mm}- & \hspace{2mm}0.50 & \hspace{4mm}0.51 & \hspace{2mm}0.48 & \hspace{4mm}0.44$^{*}$ & \hspace{2mm}0.46$^{\dagger}$ & \hspace{4mm}0.45 \\ 
\cmidrule(lr){1-11}

\multirow{1}{*}{\hspace{2mm}\makecell[l]{{ES Datasets}}}
& Llama-Reddit & - & - & \hspace{2mm}- 
& \hspace{2mm}0.30$^{*}$ & \hspace{4mm}0.10$^{*}$ 
& \hspace{2mm}0.28$^{*}$ & \hspace{4mm}0.07$^{*}$ 
& \hspace{2mm}0.29$^{*}$ & \hspace{4mm}0.09$^{*}$ \\ 
& Llama-ESConv & - & - & \hspace{2mm}- & \hspace{2mm}0.37$^{*}$ & \hspace{4mm}0.21$^{*}$ & \hspace{2mm}0.35$^{*}$ & \hspace{4mm}0.17$^{*}$ & \hspace{2mm}0.37$^{*}$ & \hspace{4mm}0.19$^{*}$ \\ 
& Llama-ExTES & - & - & \hspace{2mm}- & \hspace{2mm}0.51 & \hspace{4mm}0.54$^{*}$ & \hspace{2mm}0.50 & \hspace{4mm}0.52 & \hspace{2mm}0.48$^{\dagger}$ & \hspace{4mm}0.51 \\ 
& Llama-Psych8k & - & - & \hspace{2mm}- & \hspace{2mm}0.50 & \hspace{4mm}0.64$^{*}$ & \hspace{2mm}0.48$^{*}$ & \hspace{4mm}0.58$^{*}$ & \hspace{2mm}0.49 & \hspace{4mm}0.62$^{*}$ \\ 
\cmidrule(lr){1-11}

\multirow{1}{*}{\hspace{2mm}\makecell[l]{{ES Methods}}}
& Ask-an-Expert & - & - & \hspace{2mm}- 
& \hspace{2mm}0.34$^{*}$ & \hspace{4mm}0.18$^{*}$ 
& \hspace{2mm}0.34$^{*}$ & \hspace{4mm}0.14$^{*}$ 
& \hspace{2mm}0.32$^{*}$ & \hspace{4mm}0.15$^{*}$ \\ 
& ESCoT & - & - & \hspace{2mm}- 
& \hspace{2mm}0.28$^{*}$ & \hspace{4mm}0.06$^{*}$ 
& \hspace{2mm}0.25$^{*}$ & \hspace{4mm}0.04$^{*}$ 
& \hspace{2mm}0.25$^{*}$ & \hspace{4mm}0.05$^{*}$ \\ 
& PPDPP & - & - & \hspace{2mm}- 
& \hspace{2mm}0.47$^{*}$ & \hspace{4mm}0.37$^{*}$ 
& \hspace{2mm}0.46$^{*}$ & \hspace{4mm}0.27$^{*}$ 
& \hspace{2mm}0.44$^{*}$ & \hspace{4mm}0.31$^{*}$ \\ 
\cmidrule(lr){1-11}

\multirow{1}{*}{\hspace{2mm}\makecell[l]{{Emotion-}\\{Reinforced}}}
& SFT & - & - & \hspace{2mm}- & \hspace{2mm}0.50 & \hspace{4mm}0.51 & \hspace{2mm}0.50 & \hspace{4mm}0.43$^{*}$ & \hspace{2mm}0.49 & \hspace{4mm}0.46$^{\dagger}$ \\ 
& DPO (GPT) & 3 & 1 & \hspace{1mm}0.5 & \hspace{2mm}0.52 & \hspace{4mm}0.55$^{*}$ & \hspace{2mm}0.49 & \hspace{4mm}0.49 & \hspace{2mm}0.49 & \hspace{4mm}0.51 \\ 
& & 5 & 1 & \hspace{1mm}0.5 & \hspace{2mm}0.53$^{*}$ & \hspace{4mm}0.59$^{*}$ & \hspace{2mm}0.49$^{*}$ & \hspace{4mm}0.44$^{*}$ & \hspace{2mm}0.49 & \hspace{4mm}0.48\\ 

\cmidrule(lr){1-11}
\hspace{2mm}ES-VR (Ours) & SFT & - & - & \hspace{2mm}- & - & - & \hspace{2mm}0.48$^{\dagger}$ & \hspace{4mm}0.44$^{*}$ & \hspace{2mm}0.48$^{\dagger}$ & \hspace{4mm}0.46$^{\dagger}$ \\ 
\bottomrule
\end{tabular}%
}
\caption{ES-Value performance for models evaluated from the perspectives of seekers and supporters. The scores represent the mean win-ratio of baselines compared to our models--SFT, DPO. Statistically significant differences compared to our models are marked with * ($p$-value $< 0.05$), and differences with $p$-value $< 0.1$ are marked with $\dagger$, as determined by the Mann-Whitney U test.}
\label{table:exp2-es-value-all}
\end{table*}

\begin{table*}[ht]
\centering
\resizebox{0.4\textwidth}{!}{%
\begin{tabular}{@{}ll*{2}{l}@{}}
\toprule
\multirow{2.5}{*}{\makecell[c]{\textbf{Categories}}} & \multirow{2.5}{*}{\makecell[c]{\textbf{Models}}} & 
\multicolumn{2}{c}{\textbf{ES-Value}\textbf{$\uparrow$}} \\
\cmidrule(lr){3-4} 
& & \textbf{Seeker} & \textbf{Supporter} \\ \midrule

\multirow{1}{*}{\makecell{{Prompt-Based}}} &
GPT &
\hspace{2mm}0.46$^{*}$ & \hspace{4mm}0.28$^{*}$ \\
& Llama &
\hspace{2mm}0.48 & \hspace{4mm}0.42$^{*}$ \\\midrule

\multirow{1}{*}{\makecell[l]{{ES Datasets}}} &
Llama-Reddit & 
\hspace{2mm}0.24$^{*}$ & \hspace{4mm}0.04$^{*}$ \\
& Llama-ESConv &
\hspace{2mm}0.30$^{*}$ & \hspace{4mm}0.11$^{*}$ \\
& Llama-ExTES &
\hspace{2mm}0.44$^{*}$ & \hspace{4mm}0.37$^{*}$ \\
& Llama-Psych8k &
\hspace{2mm}0.46$^{*}$ & \hspace{4mm}0.45$^{*}$ \\\midrule

\multirow{1}{*}{\makecell[l]{{ES Methods}}} &
Ask-an-Expert &
\hspace{2mm}0.28$^{*}$ & \hspace{4mm}0.08$^{*}$ \\
& ESCoT & 
\hspace{2mm}0.24$^{*}$ & \hspace{4mm}0.03$^{*}$ \\
& PPDPP &
\hspace{2mm}0.40$^{*}$ & \hspace{4mm}0.20$^{*}$ \\\midrule

\multirow{1}{*}{\makecell{{Ours}}} &
DPO (Cactus) &
\hspace{5mm}- & \hspace{7mm}- \\
\bottomrule

\end{tabular}%
}
\caption{Comparison of models based on ES-Value---the win ratio of each model against \textit{DPO (Cactus)}. A value below 0.5 indicates that \textit{DPO (Cactus)} outperformed the baseline models, while a value above 0.5 suggests that the baselines performed better.
Statistically significant differences compared to \textit{DPO (Cactus)} are marked with * ($p$-value $< 0.05$), and differences with $p$-value $< 0.1$ are marked with $\dagger$, as determined by the Mann-Whitney U test.}
\label{table:cactus-es-value}
\end{table*}

\begin{table*}[t]
\centering
\resizebox{0.9\textwidth}{!}{%
\begin{tabular}{@{}l*{10}{l}l@{}}
\toprule
\multirow{2.5}{*}{\makecell[c]{\textbf{Models}}} & 
\multicolumn{10}{c}{\textbf{ES-Skills}↑} & \multirow{2.5}{*}{\makecell[c]{\textbf{ES-}\\\textbf{Intensity}↓}} \\ \cmidrule(lr){2-11}
& \textbf{Iden.} & \textbf{Comf.} & \textbf{Sugg.} & \textbf{Expe.} & \textbf{Info.} & \textbf{Cons.} & \textbf{Role.} & \textbf{Expr.} & \textbf{Huma.} & \textbf{Over.} & \\ \midrule

SFT & \hspace{1mm}\textbf{4.85} & \hspace{1mm}\textbf{4.90} & \hspace{1mm}\textbf{4.72} & \hspace{1mm}\textbf{3.76} & \hspace{1mm}\textbf{4.56} & \hspace{1mm}\textbf{4.99} & \hspace{1mm}\textbf{5.00} & \hspace{1mm}\textbf{4.25} & \hspace{1mm}\textbf{4.73} & \hspace{1mm}\textbf{4.80} & \hspace{4mm}\textbf{1.86} \\
- Next strategy & \hspace{1mm}4.81& \hspace{1mm}4.89 & \hspace{1mm}4.57$^{*}$ & \hspace{1mm}2.79$^{*}$ & \hspace{1mm}4.43$^{*}$ & \hspace{1mm}\textbf{4.99} & \hspace{1mm}\textbf{5.00} & \hspace{1mm}4.05$^{*}$ & \hspace{1mm}4.55$^{*}$ & \hspace{1mm}4.68$^{*}$ & \hspace{4mm}1.89 \\
- Others' experience & \hspace{1mm}4.76 & \hspace{1mm}4.88 & \hspace{1mm}4.57$^{*}$ & \hspace{1mm}3.36$^{*}$ & \hspace{1mm}4.43$^{*}$ & \hspace{1mm}4.98 & \hspace{1mm}\textbf{5.00} & \hspace{1mm}4.04$^{*}$ & \hspace{1mm}4.63$^{*}$ & \hspace{1mm}4.70$^{*}$ & \hspace{4mm}1.89 \\ \midrule

DPO (GPT) & \hspace{1mm}\textbf{4.91} & \hspace{1mm}\textbf{4.93} & \hspace{1mm}\textbf{4.78} & \hspace{1mm}\textbf{3.61} & \hspace{1mm}\textbf{4.65} & \hspace{1mm}4.99 & \hspace{1mm}\textbf{5.00} & \hspace{1mm}\textbf{4.23} & \hspace{1mm}\textbf{4.72} & \hspace{1mm}\textbf{4.85} & \hspace{4mm}1.94 \\
- Next strategy & \hspace{1mm}4.88 & \hspace{1mm}4.92 & \hspace{1mm}4.74 & \hspace{1mm}2.87$^{*}$ & \hspace{1mm}4.57$^{\dagger}$ & \hspace{1mm}\textbf{5.00} & \hspace{1mm}\textbf{5.00} & \hspace{1mm}4.08$^{*}$ & \hspace{1mm}4.59$^{*}$ & \hspace{1mm}4.77$^{*}$ & \hspace{4mm}1.90 \\
- Others' experience & \hspace{1mm}4.85 & \hspace{1mm}4.91 & \hspace{1mm}4.70$^{\dagger}$ & \hspace{1mm}3.28$^{*}$ & \hspace{1mm}4.62 & \hspace{1mm}4.99 & \hspace{1mm}\textbf{5.00} & \hspace{1mm}4.11$^{*}$ & \hspace{1mm}4.67$^{\dagger}$ & \hspace{1mm}4.80 & \hspace{4mm}\textbf{1.82} \\ \midrule

DPO (SFT) & \hspace{1mm}\textbf{4.90} & \hspace{1mm}\textbf{4.95} & \hspace{1mm}\textbf{4.80} & \hspace{1mm}\textbf{3.85} & \hspace{1mm}\textbf{4.69} & \hspace{1mm}\textbf{5.00} & \hspace{1mm}\textbf{5.00} & \hspace{1mm}\textbf{4.30} & \hspace{1mm}\textbf{4.77} & \hspace{1mm}\textbf{4.87} & \hspace{4mm}\textbf{1.75} \\
- Next strategy & \hspace{1mm}4.81$^{*}$ & \hspace{1mm}4.91 & \hspace{1mm}4.60$^{*}$ & \hspace{1mm}2.90$^{*}$ & \hspace{1mm}4.43$^{*}$ & 4.99 & \hspace{1mm}\textbf{5.00} & \hspace{1mm}4.06$^{*}$ & \hspace{1mm}4.55$^{*}$ & \hspace{1mm}4.70$^{*}$ & \hspace{4mm}1.92$^{*}$ \\
- Others' experience & \hspace{1mm}4.80$^{*}$ & \hspace{1mm}4.91 & \hspace{1mm}4.56$^{*}$ & \hspace{1mm}3.39$^{*}$ & \hspace{1mm}4.47$^{*}$ & \hspace{1mm}4.99 & \hspace{1mm}\textbf{5.00} & \hspace{1mm}4.07$^{*}$ & \hspace{1mm}4.65$^{*}$ & \hspace{1mm}4.70$^{*}$ & \hspace{4mm}1.98$^{*}$ \\ \bottomrule

\end{tabular}%
}
\caption{Comparison of ES-Skills and ES-Intensity performance based on the inclusion of self-disclosure. Statistically significant differences compared to our models are marked with * ($p$-value $< 0.05$), and differences with $p$-value $< 0.1$ are marked with $\dagger$, as determined by the Mann-Whitney U test.}
\label{table:exp3-no-sd1}
\end{table*}

\begin{table*}[t]
\centering
\resizebox{0.7\textwidth}{!}{%
\begin{tabular}{@{}ll*{5}{l}@{}}
\toprule
\multirow{2.5}{*}{\makecell[c]{\textbf{Models}}} & \multicolumn{2}{c}{\textbf{SFT}} & \multicolumn{2}{c}{\textbf{DPO (GPT)}} & \multicolumn{2}{c}{\textbf{DPO (SFT)}} \\ 
\cmidrule(lr){2-3} \cmidrule(lr){4-5} \cmidrule(lr){6-7}
& \textbf{Seeker} & \textbf{Supporter} & \textbf{Seeker} & \textbf{Supporter} & \textbf{Seeker} & \textbf{Supporter} \\ 
\midrule
- Next strategy & \hspace{2mm}0.50& \hspace{4mm}0.52 & \hspace{2mm}0.50 & \hspace{4mm}0.53$^{\dagger}$ & \hspace{2mm}0.48$^{*}$ & \hspace{4mm}0.45$^{*}$ \\
- Others' experience & \hspace{2mm}0.50 & \hspace{4mm}0.50 & \hspace{2mm}0.50 & \hspace{4mm}0.50 & \hspace{2mm}0.48$^{\dagger}$ & \hspace{4mm}0.47 \\
\bottomrule
\end{tabular}%
}
\caption{ES-Value performance for models evaluated from the perspectives of seekers and supporters, considering the impact of self-disclosure on performance. The scores represent the mean win-ratio of baselines compared to our models. Statistically significant differences compared to our models are marked with * ($p$-value $< 0.05$), and differences with $p$-value $< 0.1$ are marked with $\dagger$, as determined by the Mann-Whitney U test.}
\label{table:exp3-no-sd2}
\end{table*}

\begin{table*}[t]
\centering
\renewcommand{\arraystretch}{1.3} 
\setlength{\tabcolsep}{6pt} 
\resizebox{0.8\textwidth}{!}{%
\begin{tabular}{@{\hskip 10pt}p{4cm}<{\centering}@{\hskip 20pt}p{10cm}@{\hskip 10pt}}
\toprule
\multicolumn{1}{c}{\textbf{Category}} & {\textbf{Subcategory}} \\ \midrule
Romantic Relationship Challenges & 
Breakups or divorce \newline 
Starting a romantic relationship \newline 
Challenges in establishing a marriage \newline 
Communication difficulties in relationships \\ \midrule

Family Dynamics and Conflicts & 
Financial issues within the family \newline 
Sibling rivalry or family disputes \newline 
Challenges in parenthood and parenting \newline 
Coping with loss or grief of a family member \\ \midrule

Friendship and Interpersonal Challenges & 
Difficulty adapting to new social environments \newline 
Challenges in maintaining friendships \newline 
Conflicts with friends \\ \midrule

Career and Work-Related Challenges & 
Work-related stress and burnout \newline 
Job loss or career setbacks \newline 
Adjusting to a new job or role \newline 
Concerns about salary and bonuses \newline 
Dissatisfaction with current job \newline 
Stress related to unemployment \newline 
Ongoing depression \\ \midrule

Academic and Educational Stress & 
Dissatisfaction with current school or major \newline 
Concerns about academic performance \newline 
Stress related to studies \newline 
Difficulty entering higher education \newline 
Lack or excess of motivation to study \\ \midrule

Self-Esteem, Identity, and Personal Growth & 
Issues with self-esteem and confidence \newline 
Searching for meaning and purpose in life \newline 
Cultural identity and sense of belonging \newline 
Concerns about body image \\ \bottomrule
\end{tabular}%
}
\caption{Overview of seekers' problem categories and subcategories.}
\label{table:seeker-problems}
\end{table*}

\begin{table*}[t]
\centering
\resizebox{\textwidth}{!}{%
\begin{tabular}{@{}p{2cm}p{16cm}@{}}
\toprule
\textbf{\hspace{2mm}System} & Generate appropriate situations that require emotional support, using the given topic and value information. \\ \midrule
\textbf{\hspace{2mm}User} & 1. Emotional support topic: \textit{\{Problem category\}} \newline
- \textit{\{Subcategory 1\}} \newline
- \textit{\{Subcategory 2\}} \newline
- \textit{\{Subcategory 3\}} \newline

2. Supported value: \textit{\{Human value\}} \newline
- Definition: \textit{\{Definition of the human value\}} \newline
- Contained values: \textit{\{Contained value 1\}}, \textit{\{Contained value 2\}}, \textit{\{Contained value 3\}} \newline

Define specific situations that individuals who prioritize the given human value (item 2) might face related to the presented emotional support topic (item 1). Generate a minimum of 10 and a maximum of 30 diverse and non-overlapping situations. Write from the perspective of an individual in need of emotional support, including 'I' as the subject, and be as specific as possible. Each situation should be one sentence (e.g., I just moved in this week, and it's so hard to make friends.) Do not provide any additional explanations and separate each situation with a newline character (`\textbackslash n'). \\ \bottomrule
\end{tabular}%
}
\caption{Prompts for generating seeker situations based on problem category and human value combinations.}
\label{table:persona-gen-prompt1}
\end{table*}

\begin{table*}[t]
\centering
\resizebox{\textwidth}{!}{%
\begin{tabular}{@{}p{2cm}p{16cm}@{}}
\toprule
\textbf{\hspace{2mm}System} & Evaluate how much each situation aligns with the given value. \\ \midrule
\textbf{\hspace{2mm}User} & 1. Situations: \textit{\{Generated situations\}} \newline

2. Supported value: \textit{\{Human value\}} \newline
- Definition: \textit{\{Definition of the human value\}} \newline
- Contained values: \textit{\{Contained value 1\}}, \textit{\{Contained value 2\}}, \textit{\{Contained value 3\}} \newline

Rate the alignment of each situation with the given value on a scale of 1-5, using the criteria below to guide your assessment: \newline
- 1: The situation does not reflect any connection to the given value. The individual's concerns or actions are entirely unrelated to the principles of this value. \newline
- 2: The situation has a minimal or indirect connection to the value. It suggests the presence of the value but lacks a clear emphasis or relevance. \newline
- 3: The situation shows some aspects of the value but not as a central theme. The value is present, but other priorities seem equally important. \newline
- 4: The situation directly relates to the principles of the value, showing clear prioritization. The value significantly shapes the individual’s thoughts or actions. \newline
- 5: The situation is driven almost entirely by the given value. The value is a central, explicit factor in shaping the individual’s perspective and decisions. \newline

For each situation, provide a brief reasoning for your rating based on these criteria, and then assign the numerical rating. Provide your response in the following format: \newline
situation: (Rewrite each situation) \newline
- Reasoning: (Your explanation here) \newline
- Rating: (1-5) \\ \bottomrule
\end{tabular}%
}
\caption{Evaluation prompts for assessing alignment between situations and provided values.}
\label{table:persona-gen-prompt2}
\end{table*}

\begin{table*}[t]
\centering
\resizebox{\textwidth}{!}{%
\begin{tabular}{@{}p{3cm}p{14cm}@{}}
\toprule
\textbf{\hspace{2mm}Category} & \textbf{Details} \\ \midrule
\textbf{\hspace{2mm}Problem} & Romantic Relationship Challenges \\ 
\textbf{\hspace{2mm}Emotion} & Frustration \\ 
\textbf{\hspace{2mm}Situation} & I feel like my creativity isn't appreciated in my marriage, and it's making me question my choices. \\
\textbf{\hspace{2mm}Demographics} & Age: 30s / Gender: Female / Occupation: Designer \\ \midrule

\textbf{\hspace{2mm}Problem} & Friendship and Interpersonal Challenges \\ 
\textbf{\hspace{2mm}Emotion} & Frustration \\ 
\textbf{\hspace{2mm}Situation} & I often change hobbies and interests, but I've noticed this makes it difficult to maintain deep connections with my friends. \\
\textbf{\hspace{2mm}Demographics} & Age: 20s / Gender: Male / Occupation: College Student \\ \midrule

\textbf{\hspace{2mm}Problem} & Academic and Educational Stress \\ 
\textbf{\hspace{2mm}Emotion} & Fear \\ 
\textbf{\hspace{2mm}Situation} & I feel torn because although I want to succeed, the fear of failure is paralyzing my ability to take risks in my studies. \\
\textbf{\hspace{2mm}Demographics} & Age: 20s / Gender: Female / Occupation: College Student \\ \midrule

\textbf{\hspace{2mm}Problem} & Career and Work-Related Challenges \\ 
\textbf{\hspace{2mm}Emotion} & Anxiety \\ 
\textbf{\hspace{2mm}Situation} & I've been unemployed for months now, and the financial strain is causing me significant stress and anxiety about maintaining a comfortable lifestyle. \\
\textbf{\hspace{2mm}Demographics} & Age: 30s / Gender: Male / Occupation: Retail Manager \\ \midrule

\textbf{\hspace{2mm}Problem} & Family Dynamics and Conflicts \\ 
\textbf{\hspace{2mm}Emotion} & Anger \\ 
\textbf{\hspace{2mm}Situation} & I just set a boundary to maintain a separation between personal and financial issues, but family members keep crossing it. \\
\textbf{\hspace{2mm}Demographics} & Age: 30s / Gender: Male / Occupation: Software Developer \\ \midrule

\textbf{\hspace{2mm}Problem} & Self-Esteem, Identity, and Personal Growth \\ 
\textbf{\hspace{2mm}Emotion} & Fear \\ 
\textbf{\hspace{2mm}Situation} & I have maintained an image of success, but I'm scared of failing and letting people see my vulnerabilities. \\
\textbf{\hspace{2mm}Demographics} & Age: 40s / Gender: Female / Occupation: Entrepreneur \\ \bottomrule
\end{tabular}%
}
\caption{Examples of generated personas.}
\label{table:persona-examples}
\end{table*}

\begin{table*}[t]
\centering
\resizebox{\textwidth}{!}{%
\begin{tabular}{@{}p{2cm}p{16cm}@{}}
\toprule
\textbf{\hspace{2mm}System} & In the following conversations, you will play the role of a patient seeking help from a therapist due to emotional difficulties. Your emotional distress stems from \textit{\{Problem category\}} and the emotion you're feeling is \textit{\{Emotion type\}}. Your detailed personal information is as follows: \newline
Age Range: \textit{\{Age range\}} \newline
Gender: \textit{\{Gender\}} \newline
Occupation: \textit{\{Occupation\}} \newline\newline
Here is an example of a conversation you can refer to: \textit{\{Example of a conversation\}} \newline\newline
When responding, use only one sentence each time. Incorporate your personal information (age range, gender, and occupation) when it seems relevant, but it is not required in every response. If you feel that you have received enough emotional support and your mood has improved, end the conversation by expressing gratitude. Then, if you think it’s appropriate to conclude the session, generate `[END]' to signify the end of the conversation. You should generate only `[END]' without saying anything else. Do not end the conversation if you still feel upset or unsettled.
\\ \midrule
\textbf{\hspace{2mm}User} & Hello, I'm here to listen. What would you like to talk about today?
\\ \midrule
\textbf{\hspace{2mm}Assistant} & \textit{\{Situation\}}
\\ \bottomrule
\end{tabular}%
}
\caption{Prompts for the seeker simulator.}
\label{table:seeker-prompt}
\end{table*}

\begin{table*}[ht]
\centering
\resizebox{0.85\textwidth}{!}{%
\begin{tabular}{@{}ccccccccccc@{}}
\toprule
\multirow{2.5}{*}{\makecell[c]{\textbf{Model}\\}} & \multirow{2.5}{*}{\makecell[c]{\textbf{Details}\\}} & \multicolumn{2}{c}{\textbf{Length}} & \multicolumn{2}{c}{\textbf{Contents}} & \multicolumn{2}{c}{\textbf{Emotions}} & \multicolumn{2}{c}{\textbf{Values}} \\ 
\cmidrule(lr){3-4} \cmidrule(lr){5-6} \cmidrule(lr){7-8} \cmidrule(lr){9-10}
& & \textbf{Avg} & \textbf{Corr}↑ & \textbf{BERT}↑ & \textbf{GPT}↑ & \textbf{V-oc}↑ & \textbf{V-reg}↑ & \textbf{Cosine}↑ & \textbf{E-dist}↓ \\ 
\midrule
\makecell[c]{GPT-4o-mini} & \makecell[c]{Zero-shot} & 18.0 & 0.377 & 0.340 & 3.864 & 0.339 & \underline{0.464} & 0.714 & \textbf{0.875} \\
& \makecell[c]{One-shot} & 17.9 & 0.371 & \underline{0.342} & \textbf{4.297} & \textbf{0.371} & \textbf{0.477} & 0.713 & \underline{0.876} \\
\cmidrule(lr){1-10}
\makecell[c]{\hspace{3mm}Llama-3-8B-Instruct} & \makecell[c]{Zero-shot} & 28.6 & 0.364 & 0.338 & 3.855 & 0.311 & 0.413 & \underline{0.813} & 1.855 \\
& \makecell[c]{One-shot} & 25.4 & 0.324 & \textbf{0.346} & \underline{3.889} & \underline{0.348} & 0.433 & \textbf{0.823} & 1.806 \\
\cmidrule(lr){1-10}
\makecell[c]{Llama-ESConv} & - & 12.5 & \underline{0.390} & 0.321 & 2.871 & 0.327 & 0.438 & \underline{0.813} & 1.792 \\
\cmidrule(lr){1-10}
\makecell[c]{Llama-ExTES} & - & 20.5 & \textbf{0.439} & 0.331 & 2.837 & 0.306 & 0.429 & 0.809 & 1.771 \\
\cmidrule(lr){1-10}
\makecell[c]{ESC-Role} & - & 13.6 & 0.376 & 0.322 & 3.355 & 0.312 & 0.438 & 0.809 & 1.749 \\ 
\bottomrule
\end{tabular}%
}
\caption{Performance comparison of models as seeker simulators on the ESConv test dataset. The zero-shot condition indicates no example dialogue in the prompt, while the one-shot condition includes one example dialogue. Emotion scores reflect the correlation with actual seeker responses. V-oc represents sentiment as seven ordinal levels, ranging from -3 (very negative) to 3 (very positive), whereas V-reg quantifies sentiment intensity on a continuous scale from 0 (most negative) to 1 (most positive). Correlation analysis was conducted using Spearman's rank correlation coefficient.}
\label{table:seeker-performance}
\end{table*}

\begin{table*}[t]
\centering
\resizebox{\textwidth}{!}{%
\begin{tabular}{@{}p{2cm}p{16cm}@{}}
\toprule
\textbf{\hspace{2mm}System} & Given a conversation between a supporter and a seeker, please assess the final emotions of the seeker after the conversation.
\\ \midrule
\textbf{\hspace{2mm}User} & Conversation: \textit{\{Dialogue\}} \newline\newline
You can only reply with one of the following sentences: \newline
- very low amount of negative emotions can be inferred \newline
- low amount of negative emotions can be inferred \newline
- moderate amount of negative emotions can be inferred \newline
- high amount of negative emotions can be inferred \newline
- extreme amount of negative emotions can be inferred \newline\newline
Question: What are the final emotions of the seeker after the conversation?
\\ \bottomrule
\end{tabular}%
}
\caption{Prompts for evaluating the seeker's final emotional intensity.}
\label{table:es-intensity-prompt}
\end{table*}

\begin{table*}[t]
\centering
\resizebox{\textwidth}{!}{%
\begin{tabular}{@{}p{2cm}p{16cm}@{}}
\toprule
\textbf{\hspace{2mm}System} & Evaluate the emotional support conversation based on the given criteria.
\\ \midrule
\textbf{\hspace{2mm}User} & 1. Human Values: \textit{\{Definition of human values\}} \newline
2. Dialogue A: \textit{\{Dialogue A\}} \newline
3. Dialogue B: \textit{\{Dialogue B\}} \newline\newline
The definitions and associated values of 20 human values are provided, along with two emotional support conversations (Dialogue A and Dialogue B) between a therapist and a patient struggling with emotional difficulties. After reading both conversations, evaluate them based on the criteria below. For each criterion, first provide reasoning for your assessment, and then determine which dialogue is better for that criterion, or if it's a tie. \newline\newline
Evaluation Criteria: \newline
1) Patient's perspective: In which dialogue were positive human values more strongly explored and reinforced in the patient through the conversation? \newline
2) Therapist's perspective: In which dialogue did the therapist more effectively help the patient in exploring and reinforcing positive human values? \newline\newline
Template: \newline
1. Reasoning: (Reasoning for the evaluation of all criteria) \newline
2. Results:\newline
1) Patient's perspective: Dialogue A, Dialogue B, or Tie \newline
2) Therapist's perspective: Dialogue A, Dialogue B or Tie 
\\ \bottomrule
\end{tabular}%
}
\caption{Prompts for evaluating the effectiveness of value reinforcement (ES-Value).}
\label{table:es-value-prompt}
\end{table*}

\begin{table*}[t]
\centering
\resizebox{\textwidth}{!}{%
\begin{tabular}{@{}p{2cm}p{16cm}@{}}
\toprule
\textbf{\hspace{2mm}System} & Given a conversation between a Therapist and a Patient, please assess whether the Patient’ emotional issue has been solved after the conversation.
\\ \midrule
\textbf{\hspace{2mm}User} & You can only reply with one of the following sentences: \newline
No, the Patient feels worse. \newline
No, the Patient feels the same. \newline
No, but the Patient feels better. \newline
Yes, the Patient’s issue has been solved. \newline
The following is a conversation about \textit{\{Emotion type\}} regarding \textit{\{Problem category\}} : \textit{\{Dialogue\}} \newline
Question: Has the Patient’s issue been solved? Answer: 
\\ \bottomrule
\end{tabular}%
}
\caption{Prompts for scoring prompts for calculating emotion scores for emotion reinforcement.}
\label{table:emo_score-prompt}
\end{table*}

\begin{table*}[t]
\centering
\resizebox{0.95\textwidth}{!}{
\begin{tabular}{@{}p{5cm}p{12cm}@{}}
\toprule
\multicolumn{1}{c}{\textbf{Value Category}} & \multicolumn{1}{c}{\textbf{Definition \& Contained Values}} \\ 
\midrule
\hspace{2mm}Self-direction: thought & 
\begin{minipage}[t]{\linewidth} 
\setlength{\itemsep}{0pt} 
\setlength{\parskip}{0pt}
\setlength{\leftmargin}{5mm} 
\begin{itemize}[leftmargin=5mm, nosep] 
    \item Definition: It is good to have own ideas and interests.
    \item Contained values: Be creative, Be curious, Have freedom of thought
\end{itemize}
\end{minipage} \\
\midrule
\hspace{2mm}Self-direction: action & 
\begin{minipage}[t]{\linewidth} 
\setlength{\itemsep}{0pt} 
\setlength{\parskip}{0pt}
\setlength{\leftmargin}{5mm} 
\begin{itemize}[leftmargin=5mm, nosep] 
    \item Definition: It is good to determine one’s own actions.
    \item Contained values: Be choosing own goals, Be independent, Have freedom of action, Have privacy
\end{itemize}
\end{minipage} \\
\midrule
\hspace{2mm}Stimulation & 
\begin{minipage}[t]{\linewidth} 
\setlength{\itemsep}{0pt} 
\setlength{\parskip}{0pt}
\setlength{\leftmargin}{5mm} 
\begin{itemize}[leftmargin=5mm, nosep] 
    \item Definition: It is good to experience excitement, novelty, and change.
    \item Contained values: Have an exciting life, Have a varied life, Be daring
\end{itemize}
\end{minipage} \\
\midrule
\hspace{2mm}Hedonism & 
\begin{minipage}[t]{\linewidth} 
\setlength{\itemsep}{0pt} 
\setlength{\parskip}{0pt}
\setlength{\leftmargin}{5mm} 
\begin{itemize}[leftmargin=5mm, nosep] 
    \item Definition: It is good to experience pleasure and sensual gratification.
    \item Contained values: Have pleasure
\end{itemize}
\end{minipage} \\
\midrule
\hspace{2mm}Achievement & 
\begin{minipage}[t]{\linewidth} 
\setlength{\itemsep}{0pt} 
\setlength{\parskip}{0pt}
\setlength{\leftmargin}{5mm} 
\begin{itemize}[leftmargin=5mm, nosep] 
    \item Definition: It is good to be successful in accordance with social norms.
    \item Contained values: Be ambitious, Have success, Be capable, Be intellectual, Be courageous
\end{itemize}
\end{minipage} \\
\midrule
\hspace{2mm} Power: dominance & 
\begin{minipage}[t]{\linewidth} 
\setlength{\itemsep}{0pt} 
\setlength{\parskip}{0pt}
\setlength{\leftmargin}{5mm} 
\begin{itemize}[leftmargin=5mm, nosep] 
    \item Definition: It is good to be in positions of control over others.
    \item Contained values: Have influence, Have the right to command
\end{itemize}
\end{minipage} \\
\midrule
\hspace{2mm}Power: resources & 
\begin{minipage}[t]{\linewidth} 
\setlength{\itemsep}{0pt} 
\setlength{\parskip}{0pt}
\setlength{\leftmargin}{5mm} 
\begin{itemize}[leftmargin=5mm, nosep] 
    \item Definition: It is good to have material possessions and social resources.
    \item Contained values: Have wealth
\end{itemize}
\end{minipage} \\
\midrule
\hspace{2mm} Face& 
\begin{minipage}[t]{\linewidth} 
\setlength{\itemsep}{0pt} 
\setlength{\parskip}{0pt}
\setlength{\leftmargin}{5mm} 
\begin{itemize}[leftmargin=5mm, nosep] 
    \item Definition: It is good to maintain one’s public image.
    \item Contained values: Have social recognition, Have a good reputation
\end{itemize}
\end{minipage} \\
\midrule
\hspace{2mm}Security: personal & 
\begin{minipage}[t]{\linewidth} 
\setlength{\itemsep}{0pt} 
\setlength{\parskip}{0pt}
\setlength{\leftmargin}{5mm} 
\begin{itemize}[leftmargin=5mm, nosep] 
    \item Definition: It is good to have a secure immediate environment.
    \item Contained values: Have a sense of belonging, Have good health, Have no debts, Be neat and tidy, Have a comfortable life
\end{itemize}
\end{minipage} \\
\midrule
\hspace{2mm}Security: societal & 
\begin{minipage}[t]{\linewidth} 
\setlength{\itemsep}{0pt} 
\setlength{\parskip}{0pt}
\setlength{\leftmargin}{5mm} 
\begin{itemize}[leftmargin=5mm, nosep] 
    \item Definition: It is good to have a secure and stable wider society.
    \item Contained values: Have a safe country, Have a stable society
\end{itemize}
\end{minipage} \\
\midrule
\hspace{2mm}Tradition & 
\begin{minipage}[t]{\linewidth} 
\setlength{\itemsep}{0pt} 
\setlength{\parskip}{0pt}
\setlength{\leftmargin}{5mm} 
\begin{itemize}[leftmargin=5mm, nosep] 
    \item Definition: It is good to maintain cultural, family, or religious traditions.
    \item Contained values: Be respecting traditions, Be holding religious faith
\end{itemize}
\end{minipage} \\
\midrule
\hspace{2mm}Conformity: rules & 
\begin{minipage}[t]{\linewidth} 
\setlength{\itemsep}{0pt} 
\setlength{\parskip}{0pt}
\setlength{\leftmargin}{5mm} 
\begin{itemize}[leftmargin=5mm, nosep] 
    \item Definition: It is good to comply with rules, laws, and formal obligations.
    \item Contained values: Be compliant, Be self-disciplined, Be behaving properly
\end{itemize}
\end{minipage} \\
\midrule
\hspace{2mm}Conformity: interpersonal & 
\begin{minipage}[t]{\linewidth} 
\setlength{\itemsep}{0pt} 
\setlength{\parskip}{0pt}
\setlength{\leftmargin}{5mm} 
\begin{itemize}[leftmargin=5mm, nosep] 
    \item Definition: It is good to avoid upsetting or harming others.
    \item Contained values: Be polite, Be honoring elders
\end{itemize}
\end{minipage} \\
\midrule
\hspace{2mm}Humility & 
\begin{minipage}[t]{\linewidth} 
\setlength{\itemsep}{0pt} 
\setlength{\parskip}{0pt}
\setlength{\leftmargin}{5mm} 
\begin{itemize}[leftmargin=5mm, nosep] 
    \item Definition: It is good to recognize one’s own insignificance in the larger scheme of things.
    \item Contained values: Be humble, Have life accepted as is
\end{itemize}
\end{minipage} \\
\midrule
\hspace{2mm}Benevolence: caring & 
\begin{minipage}[t]{\linewidth} 
\setlength{\itemsep}{0pt} 
\setlength{\parskip}{0pt}
\setlength{\leftmargin}{5mm} 
\begin{itemize}[leftmargin=5mm, nosep] 
    \item Definition: It is good to work for the welfare of one’s group’s members.
    \item Contained values: Be helpful, Be honest, Be forgiving, Have the own family secured, Be loving
\end{itemize}
\end{minipage} \\
\midrule
\hspace{2mm}Benevolence: dependability & 
\begin{minipage}[t]{\linewidth} 
\setlength{\itemsep}{0pt} 
\setlength{\parskip}{0pt}
\setlength{\leftmargin}{5mm} 
\begin{itemize}[leftmargin=5mm, nosep] 
    \item Definition: It is good to be a reliable and trustworthy member of one’s group.
    \item Contained values: Be responsible, Have loyalty towards friends
\end{itemize}
\end{minipage} \\
\midrule
\hspace{2mm}Universalism: concern & 
\begin{minipage}[t]{\linewidth} 
\setlength{\itemsep}{0pt} 
\setlength{\parskip}{0pt}
\setlength{\leftmargin}{5mm} 
\begin{itemize}[leftmargin=5mm, nosep] 
    \item Definition: It is good to strive for equality, justice, and protection for all people.
    \item Contained values: Have equality, Be just, Have a world at peace
\end{itemize}
\end{minipage} \\
\midrule
\hspace{2mm}Universalism: nature & 
\begin{minipage}[t]{\linewidth} 
\setlength{\itemsep}{0pt} 
\setlength{\parskip}{0pt}
\setlength{\leftmargin}{5mm} 
\begin{itemize}[leftmargin=5mm, nosep] 
    \item Definition: It is good to preserve the natural environment.
    \item Contained values: Be protecting the environment, Have harmony with nature, Have a world of beauty
\end{itemize}
\end{minipage} \\
\midrule
\hspace{2mm}Universalism: tolerance & 
\begin{minipage}[t]{\linewidth} 
\setlength{\itemsep}{0pt} 
\setlength{\parskip}{0pt}
\setlength{\leftmargin}{5mm} 
\begin{itemize}[leftmargin=5mm, nosep] 
    \item Definition: It is good to accept and try to understand those who are different from oneself.
    \item Contained values: Be broadminded, Have the wisdom to accept others
\end{itemize}
\end{minipage} \\
\midrule
\hspace{2mm}Universalism: objectivity & 
\begin{minipage}[t]{\linewidth} 
\setlength{\itemsep}{0pt} 
\setlength{\parskip}{0pt}
\setlength{\leftmargin}{5mm} 
\begin{itemize}[leftmargin=5mm, nosep] 
    \item Definition: It is good to search for the truth and think in a rational and unbiased way
    \item Contained values: Be logical, Have an objective view
\end{itemize}
\end{minipage} \\
\bottomrule
\end{tabular}%
}
\caption{Value taxonomy introduced by \citet{kiesel2022identifying}. In this study, we focus on 20 values corresponding to the level 1 categories.}
\label{table:value-taxonomy}
\end{table*}

\begin{table*}[t]
\centering
\resizebox{0.9\textwidth}{!}{
\begin{tabular}{@{}p{11cm}p{5cm}@{}}
\toprule
\textbf{Seeker Responses} & \textbf{Detected Values} \\ 
\midrule
Yes, I accept your thought, and it gives me support. Thank you for your concern. & 
Benevolence: caring \newline 
Security: personal \newline 
Universalism: tolerance \\
\midrule
I will keep trying until I secure a new job. I will not rest. & 
Security: personal \newline 
Achievement \newline 
Self-direction: action \newline 
Power: resources \\
\midrule
That is a really valid point and is helping me see the bigger picture in life. I need to know it won’t always be this way. & 
Benevolence: caring \newline 
Security: personal \newline 
Achievement \newline 
Universalism: tolerance \newline 
Stimulation \newline 
Humility \newline 
Universalism: objectivity \\
\bottomrule
\end{tabular}
}
\caption{Examples of the seeker’s utterances in ESConv, along with the values observed in each one}
\label{table:esconv-example}
\end{table*}

\begin{table*}[t]
\centering
\renewcommand{\arraystretch}{1.3} 
\setlength{\tabcolsep}{6pt} 
\resizebox{0.8\textwidth}{!}{%
\begin{tabular}{@{\hskip 10pt}p{4cm}<{\centering}@{\hskip 20pt}p{10cm}@{\hskip 10pt}}
\toprule
\multicolumn{1}{c}{\textbf{Category}} & {\hspace{2mm}\textbf{Details}} \\ \midrule
Exploration of Issues and Challenges & 
\begin{minipage}[t]{\linewidth} 
\setlength{\itemsep}{0pt} 
\setlength{\parskip}{0pt}
\setlength{\leftmargin}{5mm} 
\begin{itemize}[leftmargin=5mm, nosep] 
    \item Insufficient understanding of the patient's key challenges and emotional struggles
    \item Lack of focus on how the patient processes emotions or responds to difficulties
\end{itemize}
\end{minipage} \\
\midrule

Exploration of Personal Interests & 
\begin{minipage}[t]{\linewidth} 
\setlength{\itemsep}{0pt} 
\setlength{\parskip}{0pt}
\setlength{\leftmargin}{5mm} 
\begin{itemize}[leftmargin=5mm, nosep] 
    \item Limited discussion on what genuinely excites or engages the patient
    \item Insufficient exploration of the patient's hobbies or areas of curiosity
    \item Lack of encouragement for the patient to share their unique interests and passions
\end{itemize}
\end{minipage} 
\\ \midrule

Exploration of Goals and Motivations & 
\begin{minipage}[t]{\linewidth} 
\setlength{\itemsep}{0pt} 
\setlength{\parskip}{0pt}
\setlength{\leftmargin}{5mm} 
\begin{itemize}[leftmargin=5mm, nosep] 
    \item Limited understanding of the patient's life goals, ambitions, and decision-making drivers
    \item Insufficient attention to articulating and clarifying meaningful objectives
\end{itemize}
\end{minipage} 
\\ \midrule

Strengths and Achievements Acknowledgment & 
\begin{minipage}[t]{\linewidth} 
\setlength{\itemsep}{0pt} 
\setlength{\parskip}{0pt}
\setlength{\leftmargin}{5mm} 
\begin{itemize}[leftmargin=5mm, nosep] 
    \item Missed opportunities to recognize the patient's unique strengths and past successes
    \item Insufficient celebration of the patient's efforts and accomplishments
    \item Limited acknowledgment of their capacity to overcome challenges and reinforce existing skills
\end{itemize}
\end{minipage} 
\\ \midrule

Emotional Resilience and Coping Strategies & 
\begin{minipage}[t]{\linewidth} 
\setlength{\itemsep}{0pt} 
\setlength{\parskip}{0pt}
\setlength{\leftmargin}{5mm} 
\begin{itemize}[leftmargin=5mm, nosep] 
    \item Insufficient guidance on building emotional resilience and adaptability
    \item Limited focus on constructive ways to navigate difficult emotions, fears, or insecurities
    \item Lack of practical approaches to manage stress, foster confidence, and maintain balance
\end{itemize}
\end{minipage} 
\\ \midrule

Focus on Achievable Goals & 
\begin{minipage}[t]{\linewidth} 
\setlength{\itemsep}{0pt} 
\setlength{\parskip}{0pt}
\setlength{\leftmargin}{5mm} 
\begin{itemize}[leftmargin=5mm, nosep] 
    \item Limited attention to setting small, manageable goals for progress
    \item Insufficient guidance on breaking down objectives into actionable tasks 
\end{itemize}
\end{minipage} 
\\ \midrule

Motivation and Alignment with Goals & 
\begin{minipage}[t]{\linewidth} 
\setlength{\itemsep}{0pt} 
\setlength{\parskip}{0pt}
\setlength{\leftmargin}{5mm} 
\begin{itemize}[leftmargin=5mm, nosep] 
    \item Missed opportunities to align goals with the patient's values and aspirations
    \item Limited encouragement for personal and professional growth opportunities
    \item Lack of suggestions for activities that resonate with the patient's interests
\end{itemize}
\end{minipage} 
\\ \midrule

Self-Compassion and Acceptance & 
\begin{minipage}[t]{\linewidth} 
\setlength{\itemsep}{0pt} 
\setlength{\parskip}{0pt}
\setlength{\leftmargin}{5mm} 
\begin{itemize}[leftmargin=5mm, nosep] 
    \item Insufficient exploration of ways to foster self-kindness and embrace imperfections
    \item Limited focus on addressing feelings of shame and building self-acceptance
\end{itemize}
\end{minipage} 
\\ \bottomrule
\end{tabular}%
}
\caption{Categories and descriptions of areas identified for improvement in value reinforcement.}
\label{table:value-issue-category}
\end{table*}

\end{document}